\documentclass[times,twocolumn,final]{elsarticle}

\usepackage{amssymb}

\usepackage{natbib}
\usepackage{hyperref}

\usepackage{amsmath,amssymb,amsfonts}
\usepackage{graphicx}
\usepackage{textcomp}

\usepackage{makecell}
\usepackage{bm}
\usepackage{array}
\usepackage{threeparttable}
\usepackage{color}
\definecolor{B}{RGB}{0,112,192}
\definecolor{G}{RGB}{0,176,80}
\definecolor{P}{RGB}{128,100,162}
\definecolor{R}{RGB}{192,0,0}
\usepackage{booktabs,diagbox,float}

\usepackage{amsmath,amssymb,amsfonts}
\usepackage{algorithmic}
\usepackage{algorithm}
\usepackage{graphicx}
\usepackage{textcomp}
\usepackage{multirow}
\usepackage{colortbl}
\usepackage{url}
\usepackage{xcolor}

\usepackage{makecell}
\usepackage{bm}
\usepackage{array}
\usepackage{threeparttable}
\usepackage{booktabs,diagbox,float}
\definecolor{g}{gray}{0.95}

\journal{Expert Systems With Applications}

\begin{document}

\begin{frontmatter}



\title{Non-Iterative Scribble-Supervised Learning with Pacing Pseudo-Masks for Medical Image Segmentation}


\author[1,2,3]{Zefan Yang}
\author[4]{Di Lin}
\author[1,2,3]{Dong Ni}
\author[1,2,3]{Yi Wang\corref{cor1}}
\cortext[cor1]{
Corresponding author: Yi Wang\\
\textit{Email addresses}:
2016222016@email.szu.edu.cn (Zefan Yang),
ande.lin1988@gmail.com (Di Lin),
nidong@szu.edu.cn (Dong Ni),
onewang@szu.edu.cn (Yi Wang)}
\address[1]{National-Regional Key Technology Engineering Laboratory for Medical Ultrasound, Guangdong Key Laboratory for Biomedical Measurements and Ultrasound Imaging, School of Biomedical Engineering, Health Science Center, Shenzhen University, Shenzhen, Guangdong, China}
\address[2]{Smart Medical Imaging, Learning and Engineering (SMILE) Lab, Shenzhen, Guangdong, China}
\address[3]{Medical UltraSound Image Computing (MUSIC) Lab, Shenzhen, Guangdong, China}
\address[4]{College of Intelligence and Computing, Tianjin University, Tianjin, China}


\begin{abstract}
Scribble-supervised medical image segmentation tackles the limitation of sparse masks.
Conventional approaches alternate between: labeling pseudo-masks and optimizing network parameters.
However, such iterative two-stage paradigm is unwieldy and could be trapped in poor local optima since the networks undesirably regress to the erroneous pseudo-masks.
To address these issues, we propose a non-iterative method where a stream of varying (pacing) pseudo-masks teach a network via consistency training, named PacingPseudo.
Our contributions are summarized as follows. First, we design a non-iterative process. This process is achieved gracefully by a siamese architecture that comparises two weight-sharing networks. The siamese architecture naturally allows a stream of pseudo-masks to assimilate a stream of predicted-masks during training.
Second, we make the consistency training effective with two necessary designs: (i) entropy regularization to obtain high-confidence pseudo-masks for effective teaching; and 
(ii) distorted augmentations to create discrepancy between the pseudo-mask and predicted-mask streams for consistency regularization.
Third, we devise a new memory bank mechanism that provides an extra source of ensemble features to complement scarce labeled pixels.
We evaluate the proposed PacingPseudo on public abdominal organ, cardiac structure, and myocardium datasets, named CHAOS T1\&T2, ACDC, and LVSC. 
Evaluation metrics include the Dice similarity coefficient (DSC) and the 95-\textit{th} percentile of Hausdorff distance (HD95).
Experimental results show that PacingPseudo achieves a 68.0\% DSC and 14.1mm HD95 on CHAOS T1, 73.7\% DSC and 12.2mm HD95 on CHAOS T2, 82.9\% DSC and 4.3mm HD95 on ACDC, and 61.4\% DSC and 11.9mm HD95 on LVSC.
These results improve the baseline method by $\geq$3.1\% in DSC and $\geq$14.2mm in HD95. These results also outcompete previous methods.
The fully-supervised method attains a 67.0\% DSC and 16.7mm HD95 on CHAOS T1, 71.2\% DSC and 12.6mm HD95 on CHAOS T2, 84.0\% DSC and 3.9mm HD95 on ACDC, and 72.9\% DSC and 7.6mm HD95 on LVSC.
PacingPseudo's performance is comparable to the fully-supervised method on CHAOS T1\&T2 and ACDC. 
Overall, the above results demonstrate the feasibility of PacingPseudo for the challenging scribble-supervised segmentation tasks.
\emph{The source code is publicly available at \url{https://github.com/zefanyang/pacingpseudo}}.
\end{abstract}




\end{frontmatter}



\section{Introduction}

\begin{figure*}[t]
	\centering
	\includegraphics[width=\textwidth]{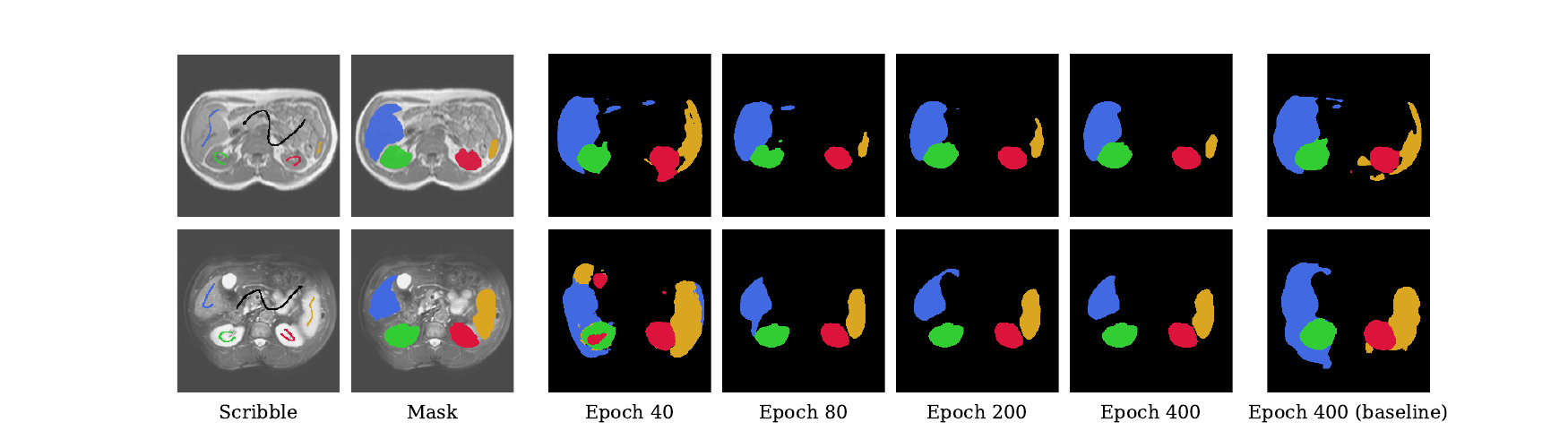}
	\caption{
		Two examples showing the evolution of training-time inference predictions.
		Left (two columns): scribble annotations and ground-truth masks.
		Middle (four columns): the predictions of our PacingPseudo updating over epochs.
		Right (one column): the predictions in the final epoch (epoch 400) of the baseline (i.e., a network trained by a partial cross-entropy loss).
		It can be observed that the predictions of PacingPseudo gradually approximate the ground-truth masks while those of the baseline present inaccuracies.
		The images are from the training sets and training is supervised solely by scribbles.}
	\label{fig:opening}
\end{figure*}

The success of deep learning in semantic segmentation still relies on great amounts of fully annotated masks~\citep{litjens2017survey, shen2017deep, isensee2021nnu, han2023hwa, uslu2023tms, qi2022directional}.
Annotating the segmentation masks inflicts high cost in the field of medical imaging because of the expertise and laborious workload needed in the process.
Scribble-supervised medical image segmentation, which trains networks supervised by scribble annotations only, can be a feasible way to reduce that burden.
Created by dragging a cursor inside target regions, scribbles are flexible to edit structures~\citep{tajbakhsh2020embracing}, but could only provide sparse labeled pixels while leaving vast regions unlabeled, posing a primary challenge in algorithm design.

Conventional scribble-supervised segmentation approaches \citep{lin2016scribblesup, can2018learning} iterate between two stages: \textit{labeling} pseudo-masks and \textit{optimizing} network parameters;
with the masks fixed, optimize the parameters, and vice versa.
However, such paradigm has two major drawbacks. Firstly, it could be trapped in poor local optima, due to the reason that the networks probably regress to errors in the initial pseudo-masks and are unable to considerably reduce such mistakes in later iterations.
Secondly, it is unwieldy especially when applied on large datasets.
To bypass the iterative process, recent studies have attained a non-iterative one.
These non-iterative approaches, which use either a regularizer \citep{tang2018normalized, tang2018regularized} or knowledge from full masks \citep{valvano2021learning} or mixed pseudo-masks \citep{zhang2022cyclemix, luo2022scribbleseg}, overlooked pure pseudo-masks, as opposed to those artificially mixed ones, for network training.

We argue that such stream can be useful and ask: \textit{In a non-iterative method, how and to what extent, can pure pseudo-masks supervised by scribbles teach a network?}
We attempt to answer the first part of the question by means of a siamese architecture~\citep{bromley1993signature} which has two weight-sharing neural networks applied on two inputs, based on the following analysis.
\textbf{(i)} Set up a non-iterative paradigm.
This paradigm, with the siamese architecture, can be achieved by translating the iterative two-stage process into: one network generating pseudo-masks supervised by scribbles (i.e., \textit{labeling}) to assimilate the predicted-masks of the other network (i.e., \textit{optimizing}) during training.
\textbf{(ii)} Use pseudo-masks to teach a network.
The pseudo-mask, supervised by scribbles, acts to regularize network parameters via consistency regularization (a regularizer) that maximizes similarity between it and the predicted-mask.
An advantage is that these pseudo-masks are being diversified than of fixed quality, due to the continually updated network parameters which are different when mapping images at each training step.
Each image's pseudo-masks are thereby varying between epochs (``pacing'').
Fig.~\ref{fig:opening} shows the predictions of ``PacingPseudo'' gradually approximate the ground-truths 
as the network learns from more equivalently improving pseudo-masks through the training process.

To answer the second part of the question, which is about improving PacingPseudo's level of performance,
we leverage insights on pseudo-labeling and augmentation from consistency training.
Firstly, since labeled pixels are scarce in scribble-supervised segmentation, output pseudo-masks remain uncertain.
\cite{xie2020unsupervised, berthelot2019mixmatch, berthelot2019remixmatch, sohn2020fixmatch} use artificial post-processing (e.g., thresholding, sharpening, or argmax) to obtain high-confidence pseudo labels,
whereas MeanTeacher \citep{yu2019uncertainty} takes self-ensembling model's predictions as pseudo-masks.
However, we empirically find these approaches are of limited effectiveness in our task, but 
the entropy regularization~\citep{grandvalet2004semi}, that regularizes pseudo-masks end-to-end, performs satisfactorily. We then provide analysis about these findings.
Secondly, augmentation is critical as it creates discrepancy between the pseudo-mask and predicted-mask branches to enable consistency regularization.
Previous studies have promoted advanced augmentation techniques \citep{berthelot2019remixmatch, xie2020unsupervised, sohn2020fixmatch} or spatial augmentations \citep{bortsova2019semi, patel2022weakly}.
In contrast, inspired by recent findings in representation learning \citep{chen2020simple, grill2020bootstrap} where augmentation serves a similar objective to create different views of an image (a positive pair) for assimilation, 
our study investigates a composition of distorted augmentations, which can be suitable and more convenient for consistency-training-based scribble-supervised segmentation.

We benchmark PacingPseudo on three public medical image datasets:
CHAOS T1 and T2 (abdominal multi-organs)~\citep{kavur2021chaos}, ACDC (cardiac structures)~\citep{bernard2018deep}, and LVSC (myocardium)~\cite{suinesiaputra2014collaborative}.
Despite its simplicity, PacingPseudo improves the baseline by large margins and consistently outcompetes previous methods in the categories of consistency training, iterative training and non-iterative training.
In some cases, PacingPseudo achieves comparable performance with its fully-supervised counterparts using ground-truth segmentation masks.

In conclusion, we list our contributions as follows:
\begin{itemize}
	\item
    We design a non-iterative paradigm to bypass the iterative two-stage paradigm proposed by previous methods~\citep{lin2016scribblesup, can2018learning}.
	We opt for the siamese architecture that naturally does ``labeling'' and ``optimizing'' during training, allowing a stream of pseudo-masks with decreasing errors to reinforce network learning.
	
	\item
    We make pure pseudo-masks sufficient for scribble-supervised learning to avoid using redundant pseudo-mask manipulation operations introduced by previous methods~\citep{luo2022scribbleseg, zhang2022cyclemix, lee2020scribble2label}.
    We utilize entropy regularization to obtain high-confidence, accurate pseudo-masks. The pseudo-masks teach a network via consistency training.
    We use distorted augmentations to create discrepancy for consistency training.
	We further study an open question about the influence of the stop-gradient operation.
	
	\item
	We develop a memory bank mechanism, whereby an extra source of information, the ensemble of embedded labeled pixels across images, is introduced to complement scarce labeled supervision.
\end{itemize}

\section{Related Work}
\subsection{Iterative vs. Non-Iterative Weakly-Supervised Segmentation}
In this section, we revisit weakly-supervised segmentation studies from an iterative or non-iterative perspective to position our study.
Conventional iterative methods pre-process pseudo-masks for network training several times relying on different techniques. 
For instance, \cite{lin2016scribblesup, can2018learning} use graph cuts \citep{boykov2004experimental} or dense conditional random fields (DCRFs) \cite{krahenbuhl2011efficient} to refine pseudo-masks (network inference predictions);
\cite{khoreva2017simple} designs heuristic prior rules to de-noise pseudo-masks for better precision; \citep{papandreou2015weakly} incorporates background and foreground biases given weak labels via expectation-maximization steps;
\citep{roth2021going} extends extreme points to pseudo-masks to supervise network training.
Other than pre-processing, \citep{dai2015boxsup} selects a small portion of candidate masks as supervision via a cost for network training in each epoch;
\cite{zhao2018deep} uses a two-step process where a detector generates proposals to be segmented.

In addition to ours, non-iterative methods do have been proposed for weakly-supervised segmentation \citep{tang2018normalized, tang2018regularized, kervadec2019constrained, lee2020scribble2label, dolz2021teach, valvano2021learning, patel2022weakly, zhang2022cyclemix, luo2022scribbleseg}.
Some studies add a regularizer to bypass pre-processing, based on shallow approaches (e.g., graph cuts) \citep{tang2018normalized, tang2018regularized} or cardinality \cite{kervadec2019constrained}.
Another category of studies transfer knowledge from full masks during training. 
While \citep{dolz2021teach} constrains weakly-supervised predictions to be similar to fully-supervised ones, full masks train an auxiliary discriminator in \citep{valvano2021learning}.
Beyond above approaches, \citep{zhang2022cyclemix, luo2022scribbleseg} use cutout or mixed (i.e., linearly interpolated) pseudo-masks for network training.
In contrast, our study argues that pure pseudo-masks (but not those artificial mixtures) can already be effective enough to teach a network
and design our method inheriting spirits from recent consistency training.

\subsection{Consistency Training}
Two aspects have been purposefully emphasized in consistency training mechanisms: pseudo-labeling to reduce uncertainty in pseudo-masks, and 
augmentation defining the neighborhood of an image to create discrepancy to enable consistency regularization. Regarding pseudo-labeling, 
\citep{berthelot2019mixmatch, berthelot2019remixmatch, yu2019uncertainty, xie2020unsupervised, sohn2020fixmatch, zou2020pseudoseg} obtain high-confidence pseudo labels using at least one of the following artificial post-processing operations:
\textbf{(i)} thresholding: eliminates a distribution whose maximum probability is smaller than a threshold;
\textbf{(ii)} sharpening: uses a temperature to sharpen a distribution;
\textbf{(iii)} argmax: truncate a distribution to one-hot encoding.
But since none of these proves effective in our task, we seek end-to-end entropy regularization \citep{grandvalet2004semi}.
Owing to its simplicity, the entropy regularization has been investigated in medical imaging \citep{dolz2021teach, luo2022word}.
However, while \citep{dolz2021teach} reports collapse when incorporating it in point-supervised segmentation, \citep{luo2022word} only tests its efficacy upon the baseline.
In contrast, we show the entropy regularization not only generally improves over the baseline in scribble-supervised segmentation, 
but also can regularize low-entropy (i.e., high-confidence) pseudo-masks to reinforce network learning via consistency training. 

In terms of augmentation in consistency training, while \citep{xie2020unsupervised, berthelot2019remixmatch, sohn2020fixmatch} favor advanced augmentation techniques 
(e.g., RandAugment \citep{cubuk2020randaugment}, CTAugment~\citep{berthelot2019remixmatch}), some studies \citep{bortsova2019semi, li2020transformation, laradji2021weakly, patel2022weakly} 
apply a spatial augmentation and its inverse version, and impose transformation equivariance.
However, we note augmentation to obtain different views of the same image is not just required in consistency training, 
but also essential in representation learning \citep{he2020momentum, chen2020simple, grill2020bootstrap, chen2021exploring}.
Both consistency training and representation learning share an objective that different views of the same image should be similar in output space.
Inspired by this insight, our study explores distortion augmentation, recently popularized in the representation-learning community \citep{chen2020simple, grill2020bootstrap}, for scribble-supervised segmentation in medical imaging.

\begin{figure*}[t]
	\centering
	\includegraphics[width=\textwidth]{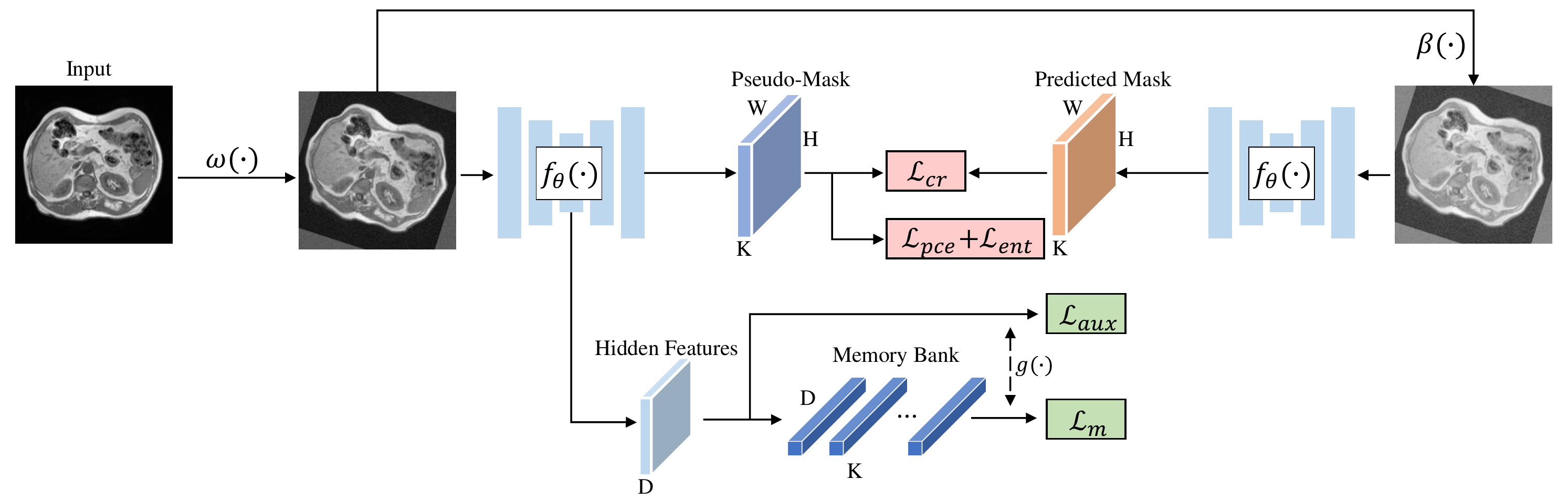}
	\caption{
		Overview of our scribble-supervised segmentation framework.
		We use a siamese architecture that contains two weight-sharing neural networks $f_\theta(\cdot)$.
		The network processes the commonly-augmented (i.e., $\omega(\cdot)$) view of the input image and produces the pseudo-mask, which is used to assimilate the predicted-mask of the further-augmented (i.e., $\beta(\cdot)$) view.
		We incorporate a memory bank, which contains the ensemble features of each semantic class mapped by a shared prediction head $g(\cdot)$ in an auxiliary path.
		The notations $\mathcal{L}_{cr}$, $\mathcal{L}_{pce}$, $\mathcal{L}_{ent}$, $\mathcal{L}_{aux}$, and $\mathcal{L}_{m}$ denote the consistency regularization loss, partial cross-entropy loss, entropy regularization loss, auxiliary loss and memory loss, respectively, which are described in details in Section~\ref{sec:method}.}
	\label{fig:framework}
\end{figure*}

\section{Method}
\label{sec:method}
Our framework (Fig.~\ref{fig:framework}) uses two weight-sharing neural networks, denoted as $f_\theta(\cdot)$.
An input image $x$ undergoes $\omega(\cdot)$ and then $\beta(\cdot)$ to produce two augmented views: a commonly-augmented view $\omega(x)$  and a further-augmented view $\beta \circ \omega(x)$.
The predictions of $\omega(x)$ serve as pseudo-masks. 
Labeled pixels in pseudo-masks are penalized by a partial cross-entropy loss $\mathcal{L}_{pce}$ described in Section \ref{sec:pce}; unlabeled pixels are regularized by an entropy regularization loss $\mathcal{L}_{ent}$ described in Section \ref{sec:ent}.
To use the pseudo-masks to guide network training, a consistency regularization loss $\mathcal{L}_{cr}$ described in Section \ref{sec:cr} maximizes similarity between the predicted-masks of $\beta \circ \omega(x)$ and the pseudo-masks.
To regularize network training, we incorporate a memory bank, an auxiliary loss $\mathcal{L}_{aux}$, and a memory loss $\mathcal{L}_{m}$ described in Section \ref{sec:mb}. The overall loss function is described in Section \ref{sec:overall loss}. The architecture of the network $f_\theta(\cdot)$ is described in Section \ref{sec:backbone}. Training details are described in Section \ref{sec:training}.

\subsection{Partial Cross-Entropy}
\label{sec:pce}
Training a network with the partial cross-entropy loss $\mathcal{L}_{pce}$ is the baseline in scribble-supervised segmentation.
The loss $\mathcal{L}_{pce}$ only penalizes the predictions of labeled pixels and ignores those of unlabeled pixels, which is written as:
\begin{equation}
\label{eq:pce}
\mathcal{L}_{pce} = \frac{1}{N} \sum_{i=0}^{N-1} \verb|CrossEntropy|(y_i, f_\theta(\omega(x))_i),
\end{equation}
where $x \in \mathbb{R}^{H\times{W}}$ denotes an input image and $N$ denotes the number of pixels in $x$.
$y \in \mathbb{R}^{H\times{W}\times{K}}$ denotes a scribble-annotated label, where $K$ is the number of classes (including the background).
$y_i \in \mathbb{R}^{K}$ is either a one-hot vector for a labeled pixel or a zero vector for an unlabeled pixel.
In such a setting, no gradient of the unlabeled pixels in $\mathcal{L}_{pce}$ is back-propagated.
$f_\theta(\omega(x))_i \in \mathbb{R}^{K}$ denotes a pre-softmax logit, where $f_\theta(\cdot)$ is a network with trainable parameters $\theta$ and $\omega(\cdot)$ is a common augmentation operation.
$\verb|CrossEntropy|(p, q)=-p \cdot \log\verb|softmax|(q)$
denotes a cross-entropy function, where $p$ and $q$ are $K$-dimensional vectors. Let $q' = \verb|softmax|(q)$ that is $q'_i = \exp(q_i)/\sum_{j=0}^{K}\exp(q_j)$ for the $i$-th channel.

\subsection{Consistency Regularization}
\label{sec:cr}
However, training the network $f_\theta(\cdot)$ with solely the partial cross-entropy loss $\mathcal{L}_{pce}$ (i.e., the baseline) is rarely satisfactory, as shown in Section~\ref{sec:fs}.
Based on our motivation of designing a non-iterative method, we use the siamese architecture to perform pseudo-mask generation and network optimization during training.
Specifically, to generate the pseudo-mask, we use the prediction of the commonly-augmented view $\tilde{y} \in \mathbb{R}^{H\times{W}\times{K}}$ that is
$\tilde{y}_i \in \mathbb{R}^{K} = \verb|softmax|(f_\theta(\omega(x))_i)$.
To perform the network optimization, we use the consistency regularization loss $\mathcal{L}_{cr}$ to maximize cross-entropy similarity between the pseudo-mask $\tilde{y}$ and the predicted-mask $f_\theta(\beta\circ\omega(x))$:
\begin{equation}
\label{eq:cr}
\mathcal{L}_{cr} = \frac{1}{N} \sum_{i=0}^{N-1} \verb|CrossEntropy|(\tilde{y}_i, f_\theta(\beta\circ\omega(x))_i),
\end{equation}
where $\beta(\cdot)$ is a further augmentation operation (described in Section~\ref{sec:augops}) that applies additional augmentation on $\omega(x)$.
In our method, we do not apply a stop-gradient operation on the pseudo-mask $\tilde{y}$ in the loss $\mathcal{L}_{cr}$, but many scribble-supervised or semi-supervised approaches did \citep{luo2022scribbleseg, xie2020unsupervised, yu2019uncertainty}.
Whether or not to apply stop-gradient leads to a different optimization trajectory, which we empirically find to considerably influence scribble-supervised segmentation.
We therefore discuss our understanding of the stop-gradient operation in Section~\ref{sec:stopgrad}.

\subsection{Entropy Regularization}
\label{sec:ent}
Our another motivation is to obtain effective low-entropy (i.e., high-confidence) pseudo-masks.
To achieve this aim, our method incorporates the entropy regularization loss $\mathcal{L}_{ent}$. 
$\mathcal{L}_{ent}$ is based on the concept of Shannon entropy. Its value reflects the degree of uncertainty in a probability distribution.
Specifically, $\mathcal{L}_{ent}$ has a low value when a distribution is nearly deterministic, but it has a high value when a distribution is nearly uniform.
The loss $\mathcal{L}_{ent}$ is written as:
\begin{equation}
\label{eq:ent}
\mathcal{L}_{ent} = - \frac{1}{N} \sum_{i=0}^{N-1} \tilde{y}_i \cdot \log \tilde{y}_i,
\end{equation}
where $\tilde{y}_i \in \mathbb{R}^{K}$ is the softmax distribution of a pseudo-mask pixel.
We train $f_\theta(\cdot)$ with the loss $\mathcal{L}_{ent}$ end-to-end, compared with \citep{yu2019uncertainty, xie2020unsupervised}
that require the artificial post-processing operations (thresholding and sharpening) to obtain low-entropy pseudo-masks to avoid degeneration.
We empirically show that the former works satisfactorily, whereas the latter are not effective. 
Related to this finding, we discuss the entropy regularization loss from a decision boundary's perspective to explain its power in Section~\ref{sec:components}.

\subsection{Memory Bank}
\label{sec:mb}
To complement scarce labeled supervision from scribbles, we introduce extra information that represents the ensemble of embedded labeled pixels across images, analogous to class prototypes in \citep{snell2017prototypical}, to regularize network learning.
Specifically, we maintains a memory bank containing the ensemble features of each semantic class, implemented as momentum moving averages of labeled-pixel features.
We build the memory bank $M \in \mathbb{R}^{K\times{D}}$ based on encoder features $f_e(\omega(x)) \in \mathbb{R}^{\frac{H}{8}\times\frac{W}{8}\times{C}}$, 
where $f_e(\cdot)$ denotes the encoder in the network $f_\theta(\cdot)$ and $C=512$ denotes the number of channels.
$f_e(\omega(x))$ are upsampled and projected to hidden features $z \in \mathbb{R}^{H\times{W}\times{D}}$, where $D=64$.
Considering a set of indices pointing to the labeled pixels of class $k$ in the label $y$ as $\mathcal{I}_k = \{i \vert y_{ik}=1, \forall i \in \{0, \cdots, N-1\}\}$, the mean features of the labeled pixels $m_k \in \mathbb{R}^D$ are computed by:
\begin{align}
m_k &= \sum_{i\in\mathcal{I}_k}s_{ik} z_i, \\
s_{ik} &= \frac{1- \texttt{sim} (M_k, z_i)}{\sum_{j\in\mathcal{I}_k}(1-\texttt{sim}(M_k, z_j))}, 
\label{eq:sim}
\end{align}
where $s_{ik}$ is a weight scalar indicating the relative importance of a pixel representation $z_i \in \mathbb{R}^D$.
Let $\verb|sim|(p, q) = p/\Vert{p}\Vert \cdot q/\Vert{q}\Vert$ denote the dot product between $\ell_2$ normalized $p$ and $q$, namely cosine similarity.
$s_{ik}$ is reversely proportional to the cosine similarity between $M_k$ and $z_i$, according to~\eqref{eq:sim}.
This prioritizes dissimilar representations for update:
\begin{equation}
M_k \leftarrow \alpha M_k + (1-\alpha) m_k,
\end{equation}
where $M_k \in \mathbb{R}^{D}$ is initialized to a zero vector (detached) and updated by the momentum moving average of $m_k$.
$\alpha=0.9$ is a momentum coefficient.
We incorporate a weight-sharing prediction head $g(\cdot)$ to map the hidden features $z$ and the memory bank $M$.
The auxiliary loss $\mathcal{L}_{aux}$ penalizes structured predictions, whilst the memory loss $\mathcal{L}_{m}$ punishes the semantic predictions derived from the memory bank features. The losses are computed by:
\begin{equation}
\mathcal{L}_{aux} = \frac{1}{N} \sum_{i=0}^{N-1} \verb|CrossEntropy|(y_i, g(z)_i),
\end{equation}
\begin{equation}
\mathcal{L}_{m} = \frac{1}{K} \sum_{k=0}^{K-1} \verb|CrossEntropy|(\hat{y}_k, g(M)_k),
\end{equation}
where $\hat{y} \in \mathbb{R}^{K\times{K}}$ (a unit matrix) represents the label of $M$.
The loss $\mathcal{L}_m$ influences the weights of $g(\cdot)$ in back-propagation, which then affects pre-softmax logits $g(z)$ from the encoder $f_e(\cdot)$ to perform regularization.

\subsection{Overall Loss}
\label{sec:overall loss}
We denote an overall loss function as:
\begin{equation}
\mathcal{L} = \mathcal{L}_{pce} + r(t)\mathcal{L}_{cr} + r(t)\mathcal{L}_{ent} + \lambda_1 \mathcal{L}_{aux} + \lambda_2 \mathcal{L}_{m},
\label{eq:overall}
\end{equation}
where $r(t)$ is an essential warm-up function applied on $\mathcal{L}_{cr}$ and $\mathcal{L}_{ent}$ to filter out noise in the early stage, and $\lambda_1$ and $\lambda_2$ are coefficients balancing loss terms.
We use an exponential form of $r(t)$ that is $r(t) = \exp(-\eta(1-t/T))$, where $t$ is an epoch index, and $T$ and $\eta$ are warm-up hyper-parameters. 
Specifically, during the first $T$ epochs, $r(t)$ ramps up from a small positive value to 1, and then remains unchanged until training is completed.
And the hyper-parameter $\eta$ controls the speed of the warm-up process.
A larger value of $\eta$ makes the warm-up process slower.
We set $T=80$ and $\eta=8$ (unless otherwise specified) which work well in our experiments.

\subsection{Backbone}
\label{sec:backbone}
\begin{figure}
	\centering
	\includegraphics[width=\columnwidth]{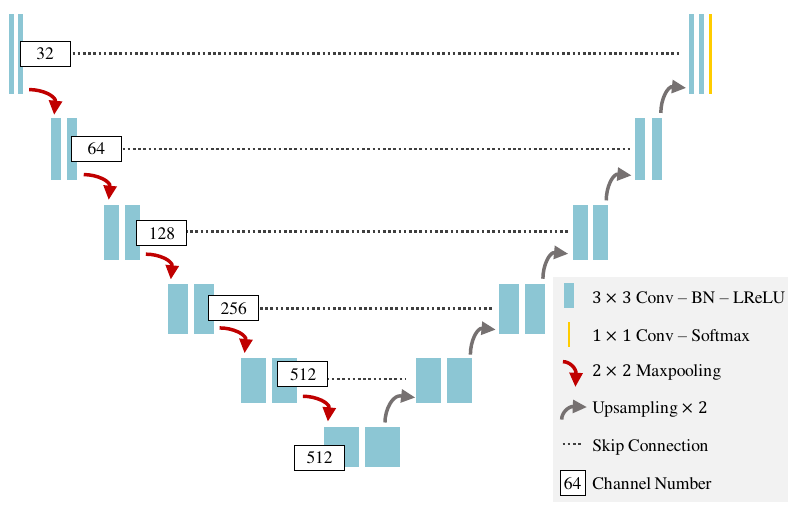}
	\caption{Network architecture. We use the encoder-decoder architecture with skip-connections as our backbone network. Each stage contains two convolutional layers in sequence. A convolutional layer comprises a 3$\times$3 convolution, batch normalization, and LeakyReLU.}
	\label{fig:net_arch}
\end{figure}

We use the UNet's encoder-decoder architecture~\citep{ronneberger2015u} as our backbone network with a few modifications, as shown in Fig.~\ref{fig:net_arch}.
We set the encoder depth (i.e., the number of encoder stages) to a value that an encoder output has a spatial size smaller than 8$\times$8, following \citep{isensee2021nnu}.
Specifically, we set the encoder depth to 6, leading to an output stride (i.e., the downsampling scale) of 32 suitable for the spatial resolutions of our datasets.
We set the initial number of channels to 32 and the maximum number of channels to 512. This means the largest channel number is set to 512 regardless of the encoder depth.
A convolutional layer is composed of a 3$\times$3 convolution, batch normalization, and LeakyReLU (with a negative slope of 0.01).
A prediction head contains a 1$\times$1 convolution and softmax operation.
We use maxpooling for downsampling and bilinear interpolation for upsampling.

\subsection{Training}
\label{sec:training}
We use Adam as our optimizer.
The Adam weight decay is 0.0003.
The Adam learning rate is 0.0001, decayed by a polynomial policy.
Specifically, the learning rate is multiplied by a scale $(1-\frac{t}{\texttt{num\_epochs}})^{0.9}$, where $t$ is an epoch index.
We use a mini-batch of 12.
On CHAOS and ACDC, we train for 400 epochs and set $T=80$ for the ramp-up function $r(t)$.
On LVSC, because of its larger scale, we train for 40 epochs and set $T=8$.
We obtain loss function's hyperparameters $\lambda_1$ and $\lambda_2$ using grid search in \{0.01, 0.1, 1\}.
Based on grid search results, we set $\lambda_1=0.01$ and $\lambda_2=1$.
We train networks on four 11 GB GPUs.
Pseudocode is presented in Algorithm \ref{tab:pseudocode}.
\emph{The source code is available at \url{https://github.com/zefanyang/pacingpseudo}.}

\begin{algorithm}[t]
\caption{Pseudocode of PacingPseudo}
\label{tab:pseudocode}

\begin{algorithmic}[1]
\REQUIRE Image $x$, scribble-annotated label $y$
\REQUIRE Backbone network $f_\theta(\cdot)$
\REQUIRE Common augmentation operation $\omega(\cdot)$
\REQUIRE Further augmentation operation $\beta(\cdot)$
\FOR{$t$ in [1, num\_epochs]}
\FOR{each minibatch}
\STATE $\backslash\backslash$ Compute pesudo-masks using $\omega(\cdot)$
\STATE $\tilde{y} = f_\theta(\omega(x))$
\STATE $\backslash\backslash$ Compute predicted-masks using $\beta\circ\omega(\cdot)$
\STATE $y' = f_\theta(\beta \circ \omega(x))$
\STATE $\backslash\backslash$ Compute memory bank features
\STATE $M_k \leftarrow \alpha M_k + (1-\alpha) m_k$
\STATE Compute $\mathcal{L}_{pce}$, $\mathcal{L}_{cr}$, $\mathcal{L}_{ent}$, $\mathcal{L}_{aux}$, $\mathcal{L}_{m}$
\STATE Update $\theta$ using Adam
\ENDFOR
\ENDFOR
\RETURN $\theta$
\end{algorithmic}
\end{algorithm}

\section{Experimental Settings}
\subsection{Datasets}
\begin{figure}[t]
	\includegraphics[width=\columnwidth]{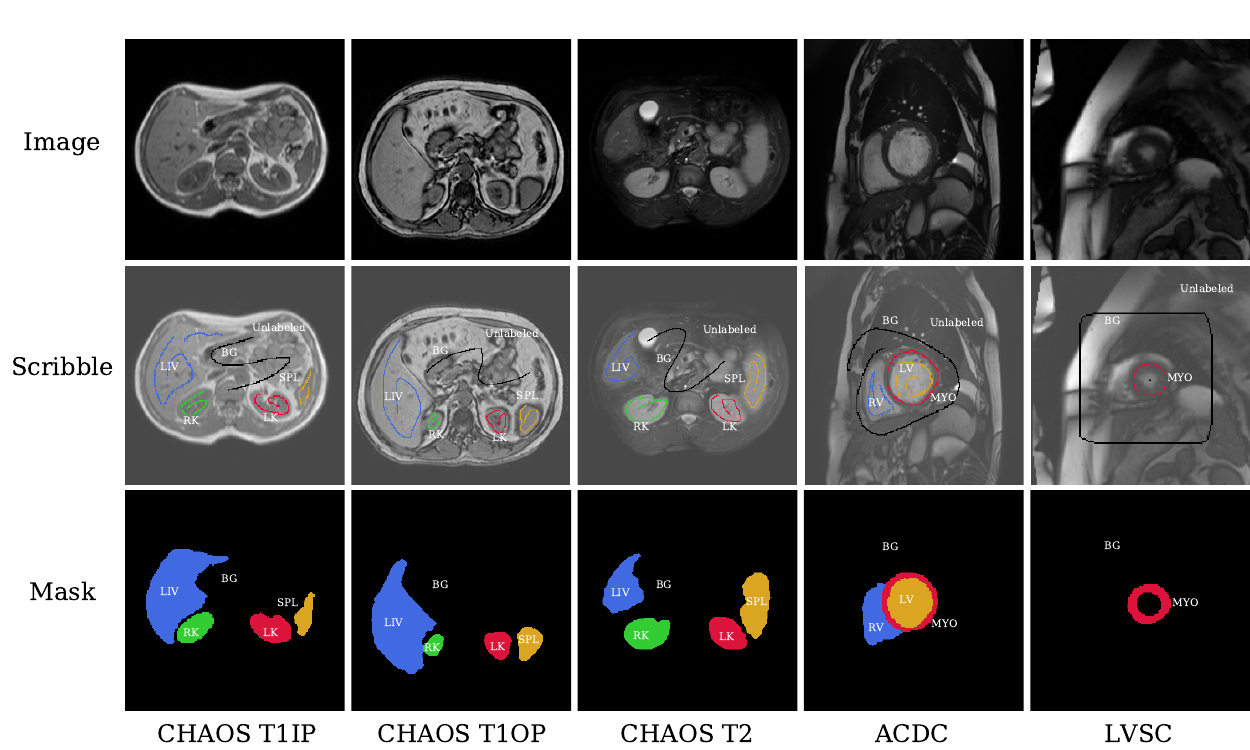}
	\centering
	\caption{Examples of images, scribbles and masks. CHAOS and ACDC use manually delineated scribbles, while
		LVSC uses artificial scribbles. T1IP and T1OP denote T1 in-phase and out-phase respectively. 
		The color-to-structure correspondences are fixed in this paper.}
	\label{fig:data}
\end{figure}

\textbf{CHAOS T1 and T2}.
The CHAOS dataset~\citep{kavur2021chaos} provides 20 patients for multi-organ segmentation.
The patients have two MRI sequences (T1-DUAL (in-phase and out-phase) and T2-SPIR) acquired by a 1.5T Phillips MRI.
T1 in-phase and out-phase are well-aligned and use a shared ground-truth, while T2-SPIR uses an independent one.
The ground-truths include four targets: liver (LIV), left kidney (LK), right kidney (RK) and spleen (SPL).
This dataset has 1,917 slices ($\sim$32 slices per sequence).
We manually delineate scribbles (including the background) based on the ground-truth using the one-pixel brush in ITK-SNAP (see Fig.~\ref{fig:data}).\footnote{\url{http://www.itksnap.org/}}
In pre-processing, we first resample images to 1.62$\times$1.62mm$^2$ (the median spacing), and then center crop or pad them to 256$\times$256 pixels (the median size).
We train separate networks for T1-DUAL (CHAOS T1) and T2-SPIR (CHAOS T2).

\textbf{ACDC}.
The ACDC dataset~\citep{bernard2018deep} provides 100 patients for heart structure segmentation.
The patients have cine-MRI sequences at the end-diastolic and end-systolic instant.
Segmentation targets include right ventricle (RV), myocardium (MYO), and left ventricle (LV).
This dataset has 1,902 slices ($\sim$10 slices per sequence).
We use scribble annotations provided by~\citep{valvano2021learning}.
We resample images to 1.51$\times$1.51mm$^2$ (the median spacing) and center crop or pad them to 224$\times$224 pixels (the median size).

\textbf{LVSC}.
The LVSC dataset~\citep{suinesiaputra2014collaborative} provides 100 patients for myocardium (MYO) segmentation.
This dataset contains ``2D+time" sequences.
We incorporate slices in short-axis view.
This dataset is relatively large, consisting of 29,086 images ($\sim$291 slices per patient).
We use artificial scribble annotations generated by skeletonizing the target and background, and then expanding the unlabeled regions.
We resample images to 1.48$\times$1.48mm$^2$ (the median spacing) and center crop or pad them to 256$\times$256 pixels (the median size).

Overall, the scribbles occupy $\sim$10$\%$ of the foreground and $\sim$0.5\% of the background on the above datasets. 
In our datasets, annotating scribbles takes 10--25 seconds per slice, whereas annotating ground-truth masks take 50--70 seconds per slice.

\subsection{Comparison Methods}
\label{sec:algdetails}
We compare PacingPseudo with three types of methods in medical imaging and computer vision.
First, to show the effectiveness of our consistency training mechanism, we compare ours with consistency training methods (UDA~\citep{xie2020unsupervised} and MeanTeacher~\citep{yu2019uncertainty}).
Second, to support our motivation of circumventing the iterative process, we compare ours with iterative training methods (NoisyStudent~\citep{xie2020self} and IterativeTraining~\citep{can2018learning}).
Third, to illustrate the advantage of our non-iterative process, we compare ours with non-iterative training methods (EntMin~\citep{luo2022word}, RegularizedLoss~\citep{tang2018regularized}, MixedPseudo~\citep{luo2022scribbleseg}, and Scribble2Label~\citep{lee2020scribble2label}).
We note, among them, IterativeTraining, EntMin, RegularizedLoss, and MixedPseudo are from the field of scribble-supervised segmentation.
Implementation details that highlight their primary difference are as follows:

\begin{itemize}
	\itemsep0em
	\item Consistency training methods
	\item[-] \textbf{UDA}: uses a siamese architecture trained with $\mathcal{L}_{pce}$ (\ref{eq:pce}) and consistency regularization.
	It eliminates distributions whose maximum probability is lower than a 0.95 threshold, and then sharpens them via a 0.5 temperature.
	
	\item[-] \textbf{MeanTeacher}: uses a teacher-student architecture trained with $\mathcal{L}_{pce}$ and consistency regularization.
	The momentum of the teacher is set to 0.999. The same thresholding and sharpening as UDA is implemented.
    
    \item Iterative training methods

	\item[-] \textbf{NoisyStudent}: adds extra noises via $\beta(\cdot)$ to images, and via dropout (with a 0.5 probability) after max-pooling at encoder stage 5 and 6 
	when training a network in iterations. Pre-processing follows the setting below.

	\item[-] \textbf{IterativeTraining}: pre-processes pseudo-masks using DCRFs in iterations.
	The hyper-parameters of DCRFs are $w_1=5$, $w_2=10$, $\sigma_\alpha=2$, $\sigma_\beta=0.1$, $\sigma_\gamma=5$.
	
	\item Non-iterative training methods
	\item[-] \textbf{EntMin}:
	trains a network with both $\mathcal{L}_{pce}$ and $\mathcal{L}_{ent}$~\eqref{eq:ent} but dose not use low-entropy pseudo-masks to reinforce network learning.
	
	\item[-] \textbf{RegularizedLoss}: proposes a DCRF regularizer.
	It first pre-trains a network with $\mathcal{L}_{pce}$, and then fine-tunes
	the network with both $\mathcal{L}_{pce}$ and the DCRF regularizer.
	
	\item[-] \textbf{MixedPseudo}: interpolates and applies argmax on predictions from two decoders to produce hard pseudo-masks, and uses them as
	supervision via the Dice loss.

        \item[-] \textbf{Scribble2Label}: uses both scribble annotations and a moving-averaged pseudo-mask to supervise network predictions during training.

\end{itemize}

\subsection{Ablation Analyses}
In Section~\ref{sec:fs}, to quantify the gap between our method and its fully-supervised counterpart, we compare the two and analyze factors that contribute to their performance difference.
In Section~\ref{sec:components}, we ablate the network components and provide our understanding of their effectiveness. 
In Section~\ref{sec:stopgrad}, we study the gradient flows of pseudo-masks in consistency regularization and discuss why no stop-gradient is better.
In Section~\ref{sec:augstrength}, we show how the further augmentation operation's strength influence performance and discuss its role in our method.
In Section~\ref{sec:scblen}, to clarify the impact of scribble coverage over targets, we evaluate the robustness of our method to different scribble lengths. 
Lastly, in Section~\ref{sec:lim}, we visualize some results to illustrate the limitations of our method.

\subsection{Evaluation Protocols}
We evaluate methods using \textit{five-fold cross-validation}.
Images are split at patient level.
We quantify the quality of predictions using two metrics:
the Dice similarity coefficient (DSC) and the 95-\emph{th} percentile of Hausdorff distance (HD95)~\citep{wang2019deep}.
While DSC (\%) measures relative overlap between the ground-truth and the prediction (higher is better),
HD95 (mm) quantifies the longest distance over the shortest distances between the surface pixels of the ground-truth and the prediction (lower is better).

\subsection{Settings of Augmentation Operations}
\label{sec:augops}
The common augmentation operation $\omega(\cdot)$ comprises geometry and noise augmentations.
Specifically, an input image is first normalized to zero mean and unit variance.
Then, it undergoes resizing, elastic deformation, rotation, horizontal and vertical flip, and Gaussian noises.
Last, the image is randomly cropped (or padded) to the original input size.
The magnitudes and probabilities of the above augmentations follow the same setting in \citep{isensee2021nnu}.

In terms of the further augmentation operation $\beta(\cdot)$, we focus on a composition of color-distorted augmentations
described in SimCLR~\citep{chen2020simple} and curate them to suit scribble-supervised segmentation in medical imaging. 
The set of operations which $\beta(\cdot)$ is sampled from is as follows:
\begin{itemize}
	\itemsep0em
	\item \textbf{Brightness}:
	increases pixel intensities by a value drawn from a uniform distribution $U(-0.8\delta, +0.8\delta)$.
	
	\item \textbf{Contrast}:
	multiplies pixel intensities by a scale drawn from $U(1-0.8\delta, 1+0.8\delta)$ and then clips them to the original min-max range.
	
	\item \textbf{GammaAugment}:
	applies a min-max normalization and then raises pixel intensities to the power of $\gamma$ drawn from $U(1-0.8\delta, 1+0.8\delta)$.
\end{itemize}
The notation $\delta \in \left(0, 1\right]$ denotes the strength of augmentations, which is set to 1 by default.
Brightness, contrast and gamma augmentation are conducted sequentially and each operation is applied with a 0.8 probability.

\section{Results and Discussion}
\label{sec:exp}
\subsection{Comparison with Previous Methods}
\label{sec:algorithms}
\subsubsection{Comparative Analysis of Quantitative Results}
As seen in Table~\ref{tab:algorithm}, PacingPseudo achieves overall best segmentation results on the three experimental datasets and
are significantly better than a majority of results of the comparison methods in both DSC and HD95 ($p$-value \textless 0.05 in two-sample t-tests).
Moreover, due to the differences in image modalities and target shapes, while the performance difference is relatively small on ACDC, 
PacingPseudo outperforms the comparison methods by large margins on CHAOS T1\&T2 and LVSC. 
Segmentation results are illustrated in Fig.~\ref{fig:methods}, 
wherein, obviously, ours are most similar to the ground-truths compared with those of the comparison methods.

\begin{table*}[t]
	\setlength{\extrarowheight}{1.2pt}
	\centering
	\addtolength{\leftskip} {-2cm}
	\addtolength{\rightskip}{-2cm}
	\caption{Comparison with previous methods.
		Results are based on five-fold cross-validation and shown in MEAN${\tiny\text{SD}}$.
		The best results are underlined, and 
		the \textbf{boldface} indicates the results are statistically different with ours ($p$ \textless~0.05).}
	\label{tab:algorithm}
	\footnotesize
	\begin{tabular}{p{0.01in}p{0.7in} p{0.2in}p{0.2in}p{0.2in}p{0.2in} p{0.2in} p{0.25in}p{0.2in}p{0.2in}p{0.2in}p{0.2in} p{0.2in}p{0.2in}p{0.2in}p{0.2in}p{0.2in}}
		\midrule[1.5pt]
		& \multicolumn{1}{l}{\multirow{2}{*}{\normalsize{Method}}} & 
		\multicolumn{5}{c}{\textbf{CHAOS T1}} & 
		\multicolumn{5}{c}{\textbf{CHAOS T2}} &
		\multicolumn{4}{c}{\textbf{ACDC}} & 
		\multicolumn{1}{c}{\textbf{LVSC}}
		\\
		
		& & 
		\multicolumn{1}{c}{LIV} & \multicolumn{1}{c}{RK} & \multicolumn{1}{c}{LK} & \multicolumn{1}{c}{SPL} & \multicolumn{1}{c}{\emph{Average}} &
		\multicolumn{1}{c}{LIV} & \multicolumn{1}{c}{RK} & \multicolumn{1}{c}{LK} & \multicolumn{1}{c}{SPL} & \multicolumn{1}{c}{\emph{Average}} &
		\multicolumn{1}{c}{RV} & \multicolumn{1}{c}{MYO} & \multicolumn{1}{c}{LV} & \multicolumn{1}{c}{\emph{Average}} &
		\multicolumn{1}{c}{MYO}
		\\
		\cmidrule(r){1-2} 
		\cmidrule(lr){3-7} \cmidrule(lr){8-12}
		\cmidrule(lr){13-16} \cmidrule(lr){17-17}
		
		\multirow{7}{*}{\rotatebox{90}{DSC}} & UDA &
		\textbf{49.0\tiny$36$} & \textbf{59.1\tiny$30$} & \textbf{51.4\tiny$34$} & \textbf{20.8\tiny$26$} & \textbf{45.1\tiny$15$} &
		\textbf{43.6\tiny$37$} & \textbf{69.2\tiny$28$} & \textbf{59.9\tiny$33$} & \textbf{25.3\tiny$34$} & \textbf{49.5\tiny$17$} & 
		\textbf{76.0\tiny$31$} & \textbf{81.5\tiny$19$} & \textbf{87.3\tiny$21$} & \textbf{81.6\tiny$05$} &
		\textbf{46.7\tiny$21$}
		\\
		
		& MeanTeacher &
		\textbf{42.8\tiny$36$} & \textbf{49.4\tiny$35$} & \textbf{51.9\tiny$36$} & \textbf{27.3\tiny$32$} & \textbf{42.9\tiny$10$} &
		\textbf{35.3\tiny$33$} & \textbf{61.8\tiny$39$} & \textbf{54.0\tiny$38$} & \textbf{30.3\tiny$37$} & \textbf{45.3\tiny$13$} &
		\textbf{68.6\tiny$33$} & \textbf{66.1\tiny$24$} & \textbf{85.7\tiny$22$} & \textbf{73.5\tiny$09$} & 
		\textbf{38.8\tiny$20$}
		\\
		
		& NoisyStudent &
		\textbf{52.0\tiny$37$} & \textbf{54.3\tiny$34$} & \textbf{55.8\tiny$34$} & \textbf{27.9\tiny$31$} & \textbf{47.5\tiny$11$} &
		\textbf{18.3\tiny$23$} & \textbf{49.5\tiny$38$} & \textbf{58.7\tiny$36$} & \textbf{30.4\tiny$35$} & \textbf{39.2\tiny$16$} & 
		\textbf{75.5\tiny$31$} & \textbf{81.4\tiny$18$} & 87.5\tiny$21$ & \textbf{81.5\tiny$05$} & 
		\textbf{41.8\tiny$18$}
		\\
		
		& IterativeTraining &
		\textbf{54.9\tiny$37$} & \textbf{56.4\tiny$33$} & \textbf{55.4\tiny$35$} & \textbf{31.7\tiny$32$} & \textbf{49.6\tiny$10$} & 
		\textbf{19.2\tiny$24$} & \textbf{51.3\tiny$38$} & \textbf{60.9\tiny$35$} & \textbf{26.8\tiny$32$} & \textbf{39.6\tiny$17$} & 
		76.5\tiny$31$ & 82.3\tiny$18$ & 87.8\tiny$20$ & 82.2\tiny$05$ & 
		\textbf{42.5\tiny$17$}
		\\
		
		& EntMin &
		\textbf{62.2\tiny$36$} & \textbf{56.1\tiny$35$} & \textbf{56.3\tiny$37$} & \textbf{40.3\tiny$35$} & \textbf{53.7\tiny$08$} & 
		\textbf{52.6\tiny$39$} & \textbf{65.3\tiny$36$} & \textbf{62.6\tiny$38$} & \textbf{39.9\tiny$39$} & \textbf{55.1\tiny$10$} &
		\textbf{75.9\tiny$30$} & 82.0\tiny$18$ & 87.8\tiny$20$ & \textbf{81.9\tiny$05$} & 
		\textbf{51.2\tiny$23$}
		\\
		
		& RegularizedLoss &
		\textbf{56.5\tiny$38$} & \textbf{52.9\tiny$34$} & \textbf{52.1\tiny$37$} & \textbf{36.4\tiny$36$} & \textbf{49.5\tiny$08$} & 
		\textbf{39.3\tiny$39$} & \textbf{24.8\tiny$37$} & \textbf{47.9\tiny$40$} & \textbf{24.3\tiny$36$} & \textbf{34.1\tiny$10$} & 
		\textbf{75.0\tiny$31$} & 82.2\tiny$18$ & 88.0\tiny$19$ & \textbf{81.8\tiny$05$} & 
		\textbf{59.4\tiny$28$}
		\\
            
		& MixedPseudo &
		\textbf{51.9\tiny$36$} & \textbf{49.8\tiny$31$} & \textbf{46.9\tiny$33$} & \textbf{22.5\tiny$28$} & \textbf{42.8\tiny$12$} & 
		\textbf{48.4\tiny$38$} & \textbf{64.5\tiny$29$} & \textbf{58.5\tiny$33$} & \textbf{21.7\tiny$30$} & \textbf{48.3\tiny$16$} & 
		75.8\tiny$31$ & 81.9\tiny$18$ & 88.3\tiny$19$ & \textbf{82.0\tiny$05$} & 
		\textbf{48.0\tiny$20$}
		\\
  
		& Scribble2Label &
		\textbf{49.7\tiny$35$} & \textbf{51.6\tiny$31$} & \textbf{47.3\tiny$34$} & \textbf{26.8\tiny$27$} & \textbf{43.9\tiny$12$} & 
		\textbf{37.5\tiny$36$} & \textbf{48.6\tiny$37$} & \textbf{53.4\tiny$35$} & \textbf{33.6\tiny$37$} & \textbf{43.3\tiny$09$} & 
		\textbf{73.6\tiny$31$} & \textbf{81.3\tiny$18$} & \textbf{87.3\tiny$21$} & \textbf{80.7\tiny$07$} & 
		\textbf{42.3\tiny$22$}
		\\
		
		& \cellcolor{g} PacingPseudo &
		\cellcolor{g} \underline{80.5\tiny$27$} & \cellcolor{g} \underline{66.5\tiny$33$} & \cellcolor{g} \underline{63.6\tiny$36$} & \cellcolor{g} \underline{61.3\tiny$36$} & \multicolumn{1}{>{\columncolor{g}}l}{\underline{68.0\tiny$07$}} &
		\cellcolor{g} \underline{79.1\tiny$30$} & \cellcolor{g} \underline{78.7\tiny$30$} & \cellcolor{g} \underline{77.1\tiny$32$} & \cellcolor{g} \underline{59.8\tiny$42$} & \multicolumn{1}{>{\columncolor{g}}l}{\underline{73.7\tiny$07$}} & 
		\cellcolor{g} \underline{77.7\tiny$29$} & \multicolumn{1}{>{\columncolor{g}}c}{\underline{82.5\tiny$17$}} & \cellcolor{g} \underline{88.4\tiny$19$} & \multicolumn{1}{>{\columncolor{g}}l}{\underline{82.9\tiny$04$}} & 
		\cellcolor{g} \underline{61.4\tiny$22$}
		\\
		\midrule
		
		\multirow{7}{*}{\rotatebox{90}{HD95}} & UDA &
		\textbf{90.1\tiny$97$} & \textbf{18.0\tiny$22$} & \textbf{19.0\tiny$18$} & \textbf{74.2\tiny$50$} & \textbf{50.3\tiny$32$} &
		\textbf{142.2\tiny$99$} & \textbf{11.3\tiny$11$} & \textbf{13.0\tiny$15$} & \textbf{39.5\tiny$26$} & \textbf{51.6\tiny$54$} &
		\textbf{8.5\tiny$17$} & \textbf{10.9\tiny$33$} & \textbf{8.3\tiny$30$} & \textbf{9.3\tiny$01$} &
		\textbf{24.5\tiny$20$}
		\\
		
		& MeanTeacher &
		\textbf{133.2\tiny$83$} & \textbf{23.0\tiny$29$} & \textbf{17.7\tiny$23$} & \textbf{50.2\tiny$41$} & \textbf{56.0\tiny$46$} &
		\textbf{185.8\tiny$66$} & \textbf{31.8\tiny$62$} & \textbf{17.2\tiny$29$} & \textbf{39.2\tiny$40$} & \textbf{68.5\tiny$68$} & 
		\textbf{49.5\tiny$67$} & \textbf{106.3\tiny$69$} & \textbf{17.5\tiny$41$} & \textbf{57.8\tiny$37$} &
		\textbf{44.8\tiny$35$}
		\\
		
		& NoisyStudent &
		\textbf{66.7\tiny$79$} & \textbf{15.0\tiny$16$} & \textbf{13.5\tiny$15$} & \textbf{45.2\tiny$33$} & \textbf{35.1\tiny$22$} & 
		\textbf{243.5\tiny$81$} & \textbf{63.2\tiny$108$} & \textbf{11.5\tiny$12$} & \textbf{55.9\tiny$72$} & \textbf{93.5\tiny$89$} & \textbf{10.1\tiny$23$} & \textbf{14.5\tiny$36$} & \textbf{8.1\tiny$24$} & \textbf{10.9\tiny$03$} &
		\textbf{29.6\tiny$21$}
		\\
		
		& IterativeTraining &
		\textbf{40.8\tiny$56$} & \textbf{14.5\tiny$12$} & \textbf{21.2\tiny$30$} & \textbf{37.8\tiny$29$} & \textbf{28.6\tiny$11$} & 
		\textbf{238.1\tiny$86$} & \textbf{53.2\tiny$94$} & \textbf{12.0\tiny$16$} & \textbf{77.1\tiny$96$} & \textbf{95.1\tiny$86$} & 
		\textbf{8.0\tiny$19$} & \textbf{9.7\tiny$27$} & \textbf{6.7\tiny$21$} & \textbf{8.1\tiny$01$} &
		\textbf{26.1\tiny$18$}
		\\
		
		& EntMin &
		\textbf{34.1\tiny$47$} & \textbf{16.5\tiny$22$} & \textbf{14.7\tiny$22$} & \textbf{29.5\tiny$26$} & \textbf{23.7\tiny$08$} & 
		\textbf{51.5\tiny$62$} & \textbf{11.8\tiny$22$} & \textbf{9.1\tiny$16$} & \textbf{24.5\tiny$33$} & \textbf{24.2\tiny$17$} &
		\textbf{7.3\tiny$12$} & \textbf{5.3\tiny$13$} & \textbf{4.1\tiny$11$} & \textbf{5.6\tiny$01$} & 
		\textbf{22.7\tiny$27$}
		\\
		
		& RegularizedLoss &
		\textbf{47.0\tiny$62$} & \textbf{17.1\tiny$19$} & \textbf{14.9\tiny$16$} & \textbf{31.8\tiny$34$} & \textbf{27.7\tiny$13$} & 
		\textbf{131.9\tiny$96$} & \textbf{136.4\tiny$93$} & \textbf{16.9\tiny$30$} & \textbf{81.9\tiny$92$} & \textbf{91.8\tiny$48$} & 
		\textbf{9.2\tiny$18$} & \textbf{6.3\tiny$18$} & \textbf{4.7\tiny$14$} & \textbf{6.7\tiny$02$} &
		\textbf{12.7\tiny$23$}
		\\

		& MixedPseudo &
		\textbf{51.9\tiny$63$} & \textbf{21.0\tiny$19$} & \textbf{19.9\tiny$17$} & \textbf{50.2\tiny$37$} & \textbf{35.8\tiny$15$} & 
		\textbf{56.8\tiny$69$} & \textbf{14.6\tiny$23$} & \textbf{15.6\tiny$28$} & \textbf{42.9\tiny$30$} & \textbf{32.5\tiny$18$} & 
		\textbf{8.1\tiny$15$} & \textbf{5.7\tiny$14$} & \textbf{4.3\tiny$12$} & \textbf{6.0\tiny$02$} & 
		\textbf{18.6\tiny$16$}
		\\

		& Scribble2Label &
		\textbf{41.7\tiny$59$} & \textbf{18.2\tiny$17$} & \textbf{15.3\tiny$14$} & \textbf{46.4\tiny$35$} & \textbf{30.4\tiny$17$} & 
		\textbf{59.7\tiny$64$} & \textbf{17.8\tiny$26$} & \textbf{16.3\tiny$30$} & \textbf{52.8\tiny$67$} & \textbf{36.7\tiny$23$} & 
		\textbf{10.2\tiny$16$} & \textbf{6.5\tiny$16$} & \textbf{7.3\tiny$13$} & \textbf{8.0\tiny$02$} & 
		\textbf{27.3\tiny$18$}
		\\

		& \cellcolor{g} PacingPseudo &
		\cellcolor{g} \underline{17.7\tiny$30$} & \cellcolor{g} \underline{11.1\tiny$18$} & \cellcolor{g} \underline{11.1\tiny$19$} & \cellcolor{g} \underline{16.6\tiny$19$} & \multicolumn{1}{>{\columncolor{g}}l}{\underline{14.1\tiny$03$}} & 
		\cellcolor{g} \underline{20.9\tiny$30$} & \cellcolor{g} \underline{5.6\tiny$09$} & \cellcolor{g} \underline{6.5\tiny$14$} & \cellcolor{g} \underline{15.7\tiny$28$} & \multicolumn{1}{>{\columncolor{g}}l}{\underline{12.2\tiny$06$}} & 
		\cellcolor{g} \underline{5.7\tiny$07$} & \multicolumn{1}{>{\columncolor{g}}l}{\underline{3.8\tiny$06$}} & \cellcolor{g} \underline{3.4\tiny$06$} & \multicolumn{1}{>{\columncolor{g}}l}{\underline{4.3\tiny$01$}} & 
		\cellcolor{g} \underline{11.9\tiny$15$}
		\\
		
		\midrule[1.5pt]
	\end{tabular}
\end{table*}


\begin{figure*}[t]
	\centering
	\addtolength{\leftskip} {-2cm}
	\addtolength{\rightskip}{-2cm}
	\includegraphics[width=\textwidth]{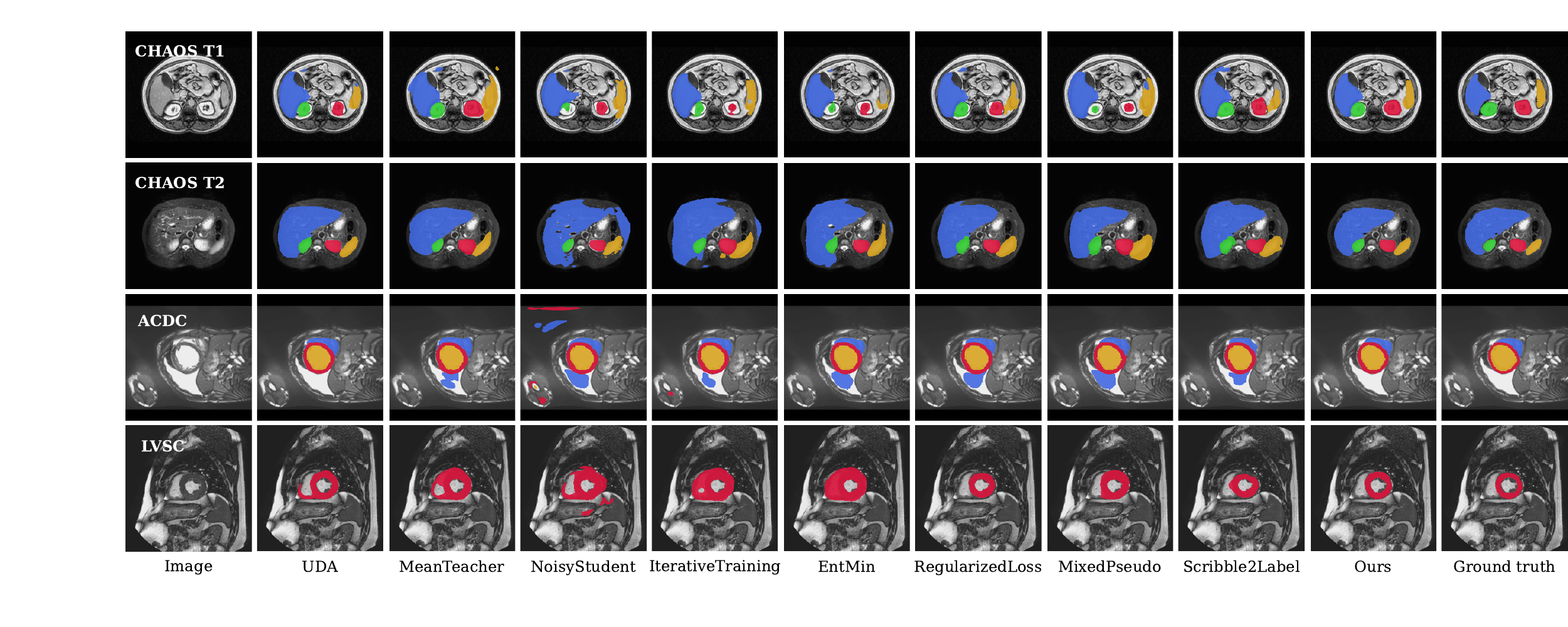}
	\caption{Qualitative comparison of the segmentations from PacingPseudo and previous methods.
		They all are supervised solely by scribble annotations.}
	\label{fig:methods}
\end{figure*}

Next, we provide analysis related to the experimental results of the comparison methods.
The main difference between UDA and PacingPseudo is that while UDA uses the artificial post-processing operations to manipulate pseudo-masks, 
PacingPseudo uses the end-to-end loss $\mathcal{L}_{ent}$~\eqref{eq:ent}.
The better performance of PacingPseudo suggests that the artificial operations are not effective for scribble-supervised segmentation.
Our explanation is that artificially processed pseudo-masks (which, for example, cover only high-confidence regions) are not what a network predicts, 
which leads to too serious inconsistency to be handled via consistency regularization and causes unsatisfactory performance.
Besides, MeanTeacher obtains the worst performance compared with UDA and PacingPseudo (both using the siamese architecture).
We conjecture two factors attributed to its failure: (i) the moving-averaged model's predictions are not quality pseudo-mask as opposed to our end-to-end regularized ones, and 
(ii) the stop-gradient operation inherently applied on pseudo-masks hampers its performance, as discussed in Section \ref{sec:stopgrad}.\footnote{The stop-gradient is applied on pseudo-masks as the teacher model simply uses moving-averaged parameters and need not be updated by gradients.}
Lastly, PacingPseudo outperforms the iterative NoisyStudent and IterativeTraining and the non-iterative EntMin, RegularizedLoss, MixedPsuedo, and Scribble2Label.
While these methods use pre-processed pseudo-masks, regularizers, or interpolated hard pseudo-masks, our method uses regularized high-confidence pseudo-masks to teach a network and shows better performance.

\subsubsection{Comparative Analysis of Computational Efficiency}
The computational efficiency of the proposed method and previous methods is significantly influenced by two factors: the network architecure and optimization process. 
The network architecture can be single or dual. The optimization process can be iterative or non-iterative. 
UDA and MeanTeacher use dual networks and non-iterative optimization as PacingPseudo does. They hence have similar computational costs. 
However, PacingPseudo performs considerably better than UDA and MeanTeacher, highlighting the superiority of our method.
NoisyStudent and IterativeTraining utilize a single network and three-step iterative process with DCRF or noising pre-processing.
The iterative process renders them unwieldy. Worse still, their performance is much worse than our PacingPseudo's.
The remainder of methods (EntMIN, RegularizedLoss, MixedPseduo, and Scribble2Label) all use a single network and non-iterative optimization.
They differ from PacingPseudo in the use of the single network architecture, which can lead to decrease in computational costs, but comes at a cost of performance loss.


\subsection{Comparison with the Baseline and Full Supervision}
\label{sec:fs}
We compare PacingPseudo with the baseline and fully-supervised counterparts.
While the baseline trains a network with the partial cross-entropy loss $\mathcal{L}_{pce}$~\eqref{eq:pce},
the fully-supervised counterparts perform network training with either $\mathcal{L}_{ce}$ or with $\mathcal{L}_{ce} + \mathcal{L}_{Dice}$ (an improved implementation),
where $\mathcal{L}_{ce}$ and $\mathcal{L}_{Dice}$ denote the cross-entropy loss and Dice loss~\citep{milletari2016v} respectively and are both based on the ground-truth segmentation masks. 

\begin{table*}[h!]
	\setlength{\extrarowheight}{1.2pt}
	\centering
	\addtolength{\leftskip} {-2cm}
	\addtolength{\rightskip}{-2cm}
	\caption{Comparison with the baseline and full supervision.
		Results are based on five-fold cross-validation and shown in MEAN${\tiny\text{SD}}$.
		The best results are underlined,
		and the \textbf{boldface} indicates the results are statistically different with ours ($p$ \textless~0.05).}
	\label{tab:fullsup}
	\footnotesize
	\begin{tabular}{p{0.01in}p{0.9in}p{0.2in}p{0.2in}p{0.2in}p{0.2in}p{0.2in}p{0.22in}p{0.2in}p{0.2in}p{0.2in}p{0.2in}p{0.2in}p{0.2in}p{0.2in}p{0.2in}p{0.2in}}
		\midrule[1.5pt]
		& \multirow{2}{*}{\normalsize{Method}} & 
		\multicolumn{5}{c}{\textbf{CHAOS T1}} & 
		\multicolumn{5}{c}{\textbf{CHAOS T2}} & 
		\multicolumn{4}{c}{\textbf{ACDC}} & 
		\multicolumn{1}{c}{\textbf{LVSC}}
		\\
		
		& & 
		\multicolumn{1}{c}{LIV} & \multicolumn{1}{c}{RK} & \multicolumn{1}{c}{LK} & \multicolumn{1}{c}{SPL} & \multicolumn{1}{c}{\emph{Average}} &
		\multicolumn{1}{c}{LIV} & \multicolumn{1}{c}{RK} & \multicolumn{1}{c}{LK} & \multicolumn{1}{c}{SPL} & \multicolumn{1}{c}{\emph{Average}} &
		\multicolumn{1}{c}{RV} & \multicolumn{1}{c}{MYO} & \multicolumn{1}{c}{LV} & \multicolumn{1}{c}{\emph{Average}} &
		\multicolumn{1}{c}{MYO}
		\\ 
		\cmidrule(r){1-2} 
		\cmidrule(lr){3-7} \cmidrule(lr){8-12}
		\cmidrule(lr){13-16} \cmidrule(lr){17-17}		
		
		\multirow{4}{*}{\rotatebox{90}{DSC}} & Baseline \tiny{$\mathcal{L}_{pce}$} &
		\textbf{45.1\tiny$37$} & \textbf{48.7\tiny$34$} & \textbf{49.5\tiny$36$} & \textbf{22.8\tiny$29$} & \textbf{41.5\tiny$11$} &
		\textbf{18.4\tiny$22$} & \textbf{51.3\tiny$37$} & \textbf{46.8\tiny$38$} & \textbf{23.9\tiny$32$} & \textbf{35.1\tiny$14$} &
		\textbf{73.1\tiny$32$} & \textbf{79.6\tiny$20$} & \textbf{86.7\tiny$21$} & \textbf{79.8\tiny$06$} &
		\textbf{36.9\tiny$18$}
		\\
		
		& Full sup. \tiny{$\mathcal{L}_{ce}$} &
		77.9\tiny$30$ & 64.4\tiny$37$ & 64.7\tiny$38$ & \textbf{53.1\tiny$40$} & \textbf{65.0\tiny$09$} &
		77.8\tiny$33$ & 75.2\tiny$35$ & \textbf{65.3\tiny$41$} & 62.8\tiny$42$ & 70.3\tiny$06$ &
		\textbf{74.8\tiny$33$} & 83.2\tiny$19$ & 88.7\tiny$20$ & 82.2\tiny$06$ & 
		\textbf{71.9\tiny$29$}
		\\
		
		& Full sup. \tiny{$\mathcal{L}_{ce}$+$\mathcal{L}_{Dice}$} &
		78.6\tiny$30$ & 65.9\tiny$37$ & \underline{67.4\tiny$39$} & \textbf{55.8\tiny$40$} & \textbf{67.0\tiny$08$} &
		77.9\tiny$34$ & 74.5\tiny$36$ & \textbf{67.6\tiny$40$} & \underline{64.9\tiny$42$} & 71.2\tiny$05$ & 
		77.4\tiny$32$ & \textbf{\underline{84.9\tiny$19$}} & \textbf{\underline{89.8\tiny$20$}} & \underline{84.0\tiny$05$} & 
		\textbf{\underline{72.0\tiny$29$}}
		\\
		
		& \cellcolor{g} PacingPseudo &
		\cellcolor{g} \underline{80.5\tiny$27$} & \cellcolor{g} \underline{66.5\tiny$33$} & \cellcolor{g} 63.6\tiny$36$ & \cellcolor{g} \underline{61.3\tiny$36$} & \multicolumn{1}{>{\columncolor{g}}l}{\underline{68.0\tiny$07$}} &
		\cellcolor{g} \underline{79.1\tiny$30$} & \cellcolor{g} \underline{78.7\tiny$30$} & \cellcolor{g} \underline{77.1\tiny$32$} & \cellcolor{g} 59.8\tiny$42$ & \multicolumn{1}{>{\columncolor{g}}l}{\underline{73.7\tiny$07$}} & 
		\cellcolor{g} \underline{77.7\tiny$29$} & \multicolumn{1}{>{\columncolor{g}}c}{82.5\tiny$17$} & \cellcolor{g} 88.4\tiny$19$ & \multicolumn{1}{>{\columncolor{g}}l}{82.9\tiny$04$} & 
		\cellcolor{g} 61.4\tiny$22$
		\\
		
		\midrule
		
		\multirow{4}{*}{\rotatebox{90}{HD95}} & Baseline \tiny{$\mathcal{L}_{pce}$} &
		\textbf{120.5\tiny$99$} & \textbf{20.2\tiny$23$} & \textbf{17.0\tiny$20$} & \textbf{57.1\tiny$44$} & \textbf{53.7\tiny$42$} & 
		\textbf{257.6\tiny$65$} & \textbf{32.7\tiny$52$} & \textbf{17.2\tiny$26$} & \textbf{49.8\tiny$38$} & \textbf{89.3\tiny$98$} & 
		\textbf{18.5\tiny$37$} & \textbf{24.5\tiny$48$} & \textbf{12.5\tiny$34$} & \textbf{18.5\tiny$05$} &
		\textbf{37.7\tiny$25$}
		\\
		
		& Full sup. \tiny{$\mathcal{L}_{ce}$} &
		\textbf{31.1\tiny$52$} & \textbf{21.0\tiny$37$} & 12.8\tiny$25$ & \textbf{21.4\tiny$28$} & \textbf{21.6\tiny$06$} & 
		20.5\tiny$33$ & \textbf{21.1\tiny$44$} & \textbf{12.8\tiny$28$} & 13.7\tiny$23$ & 17.0\tiny$04$ & 
		\textbf{7.4\tiny$14$} & 4.3\tiny$09$ & \textbf{4.0\tiny$10$} & \textbf{5.2\tiny$02$} &
		\textbf{\underline{7.4\tiny$14$}}
		\\
		
		& Full sup. \tiny{$\mathcal{L}_{ce}$+$\mathcal{L}_{Dice}$} &
		\textbf{23.4\tiny$39$} & 13.5\tiny$27$ & \textbf{\underline{8.8\tiny$19$}} & \textbf{21.1\tiny$3$} & \textbf{16.7\tiny$06$} & 
		\underline{20.0\tiny$33$} & \textbf{9.7\tiny$26$} & 8.9\tiny$19$ & \underline{11.7\tiny$19$} & 12.6\tiny$04$ &
		5.8\tiny$11$ & \textbf{\underline{3.1\tiny$05$}} & \textbf{\underline{2.9\tiny$05$}} & \textbf{\underline{3.9\tiny$01$}} & 
		\textbf{7.6\tiny$15$}
		\\
		
		& \cellcolor{g} PacingPseudo &
		\cellcolor{g} \underline{17.7\tiny$30$} & \cellcolor{g} \underline{11.1\tiny$18$} & \cellcolor{g} 11.1\tiny$19$ & \cellcolor{g} \underline{16.6\tiny$19$} & \multicolumn{1}{>{\columncolor{g}}l}{\underline{14.1\tiny$03$}} & 
		\cellcolor{g} 20.9\tiny$30$ & \cellcolor{g} \underline{5.6\tiny$09$} & \cellcolor{g} \underline{6.5\tiny$14$} & \cellcolor{g} 15.7\tiny$28$ & \multicolumn{1}{>{\columncolor{g}}l}{\underline{12.2\tiny$06$}} & 
		\cellcolor{g} \underline{5.7\tiny$07$} & \multicolumn{1}{>{\columncolor{g}}l}{3.8\tiny$06$} & \cellcolor{g} 3.4\tiny$06$ & \multicolumn{1}{>{\columncolor{g}}l}{4.3\tiny$01$} & 
		\cellcolor{g} 11.9\tiny$15$
		\\
		
		\midrule[1.5pt]
	\end{tabular}
\end{table*}

As seen in Table~\ref{tab:fullsup}, PacingPseudo improves the baseline by large margins in both DSC and HD95.
More important, PacingPseudo attains comparable performance with its fully-supervised counterparts on CHAOS T1\&T2 and ACDC.
To better understand this finding, we compare and analyze their mechanisms.
The key difference lies in the mask. 
While the full supervision can access full masks, our method uses only scribbles and leaves the remaining regions unlabeled.
But we compensate this deficiency by using regularized high-confidence pseudo-masks (as opposed to one-hot full masks) to teach a network.
Another difference is the network architecture.
While the full supervision uses a single network, our method leverages weigh-sharing networks which give a natural advantage that augmentations in the predicted-mask branch can always act to expand the training set.
These factors could have contributed to narrowing their performance gap.
Beyond the above finding, our method, however, does not catch up with its fully-supervised counterparts on LVSC.
This may be because scribble annotations provide limited boundary constraint to control the extent of the thin, ringed myocardium, which is the limitations of our method discussed in Section \ref{sec:lim}.
But, overall, PacingPseudo could be promising to bridge the gap between the scribble-supervised and fully-supervised segmentation.

\begin{table}[t]
	\centering
	\caption{Ablation analyses: network components.
		The best results are underlined, and
		the \textbf{boldface} indicates the results are statistically different with ours ($p$ \textless~0.05).
	}
	\label{tab:ablation}
	\footnotesize
	\begin{tabular}{p{0.01in}p{0.65in} p{0.5in}<{\centering}p{0.5in}<{\centering}p{0.3in}<{\centering}p{0.3in}<{\centering}}
		\midrule[1.5pt]
		& Method & \multicolumn{1}{c}{CHAOS T1} & \multicolumn{1}{c}{CHAOS T2} & \multicolumn{1}{c}{ACDC} & \multicolumn{1}{c}{LVSC} \\ 
		\cmidrule(r){1-2} 
		\cmidrule(lr){3-3} \cmidrule(lr){4-4} 
		\cmidrule(lr){5-5} \cmidrule(lr){6-6}
		
		\multirow{4}{*}{\rotatebox{90}{DSC}} & Baseline & \textbf{41.5\tiny$11$} & \textbf{35.1\tiny$14$} & \textbf{79.8\tiny$06$} & \textbf{36.9\tiny$18$} \\
		& EntMin & \textbf{53.7\tiny$08$} & \textbf{55.1\tiny$10$} & \textbf{81.9\tiny$05$} & \textbf{51.2\tiny$23$} \\
		& + Memory & \textbf{56.8\tiny$08$} & \textbf{59.6\tiny$11$} & 82.4\tiny$05$ & \textbf{54.6\tiny$22$} \\ 
		& \cellcolor{g} + Consistency & \cellcolor{g} \underline{68.0\tiny$07$} & \cellcolor{g} \underline{73.7\tiny$08$} & \cellcolor{g} \underline{82.9\tiny$04$} & \cellcolor{g} \underline{61.4\tiny$22$} \\
		\midrule
		\multirow{4}{*}{\rotatebox{90}{HD95}} & Baseline & \textbf{53.7\tiny$42$} & \textbf{89.3\tiny$98$} & \textbf{18.5\tiny$05$} & \textbf{37.7\tiny$25$} \\
		& EntMin & \textbf{23.7\tiny$08$} & \textbf{24.2\tiny$17$} & \textbf{5.6\tiny$01$} & \textbf{22.7\tiny$27$} \\
		& + Memory & \textbf{23.3\tiny$10$} & \textbf{22.6\tiny$17$} & \textbf{5.0\tiny$01$} & \textbf{14.7\tiny$14$} \\ 
		& \cellcolor{g} + Consistency & \cellcolor{g} \underline{14.1\tiny$03$} & \cellcolor{g} \underline{12.2\tiny$06$} & \cellcolor{g} \underline{4.3\tiny$01$} & \cellcolor{g} \underline{11.9\tiny$15$} \\
		\midrule[1.5pt]
	\end{tabular}
\end{table}

\subsection{Ablation: Network Components}
\label{sec:components}
We study the efficacy of the network components in this subsection.
In Table~\ref{tab:ablation}, EntMin incorporates the entropy regularization loss $\mathcal{L}_{ent}$~\eqref{eq:ent} into the baseline,
then the ``+Memory" entry implements the memory bank setting in an auxiliary path,
and the ``+Consistency" uses the siamese architecture and adds the consistency regularization loss $\mathcal{L}_{cr}$~\eqref{eq:cr}.

As shown in Table~\ref{tab:ablation}, it is surprising that simply adding the loss $\mathcal{L}_{ent}$ (the EntMin) brings considerable improvements over the baseline.
We conjecture that this is because the loss $\mathcal{L}_{pce}$ in the baseline constrains only a few pixels to be one-hot, which leads to uncertainty.
Adding the loss $\mathcal{L}_{ent}$, however, incorporates all pixels and gradually regularizes them to be confident \citep{dolz2021teach},
which in effect encourages a network to learn a beneficial low-density decision boundary \citep{chapelle2005semi} that
could separate classes clearly and hence improve validation accuracy.
Then, the ``+Memory" can further boost the performance. This supports our motivation of introducing a source of ensemble features to complement scarce supervision.
The ``+Consistency" brings another performance leaps in both metrics.
This validates the effectiveness of using a stream of regularized high-confidence pseudo-masks to reinforce network learning in a non-iterative manner (via a siamese architecture).

\begin{figure}[t]
	\includegraphics[width=\columnwidth]{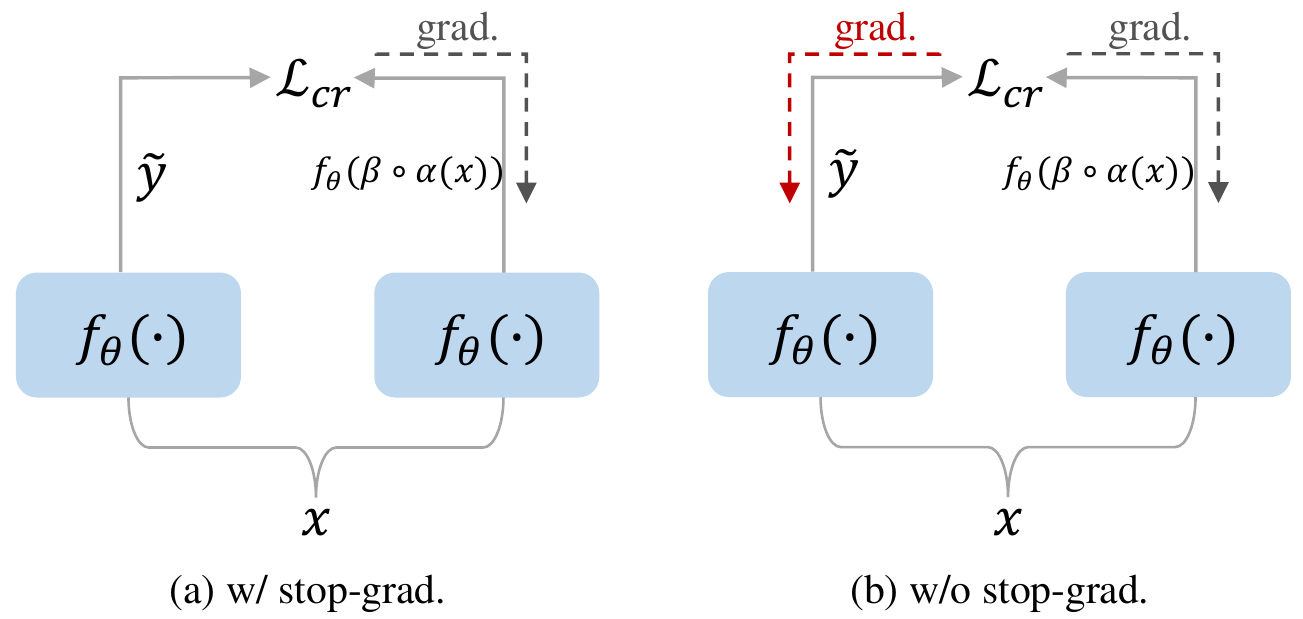}
	\centering
	\caption{Stop-gradient on Pseudo-Masks. (a) With the stop-gradient, the gradients of the pseudo-mask $\tilde{y}$ are prevented from back-propagation, 
		while (b) without the stop-gradient, the gradients of $\tilde{y}$ are back-propagated. The above framework illustrates $\mathcal{L}_{cr}$ only for clarity.}
	\label{fig:stop-grad}
\end{figure}

\subsection{Ablation: Stop-Gradient on Pseudo-Masks}
\label{sec:stopgrad}
Regarding the designs of pseudo-masks, we have discussed why artificial operations (e.g., thresholding and sharpening) are not effective in Section \ref{sec:algorithms}.
In this subsection, we investigate another question, whether or not to apply the stop-gradient operation on pseudo-masks.
Specifically, we design two settings:
one with the stop-gradient on the pseudo-mask $\tilde{y}$ in the loss $\mathcal{L}_{cr}$~\eqref{eq:cr} \citep{luo2022scribbleseg, xie2020unsupervised, yu2019uncertainty} 
and the other without the stop-gradient (ours) (see Fig. \ref{fig:stop-grad}).

\begin{table}[t]
	\centering
	\caption{Ablation analyses: with and without stop-gradient.
		The best results are underlined.}
	\label{tab:stopgrad}
	\footnotesize
	\begin{tabular}{p{0.01in}p{0.65in} p{0.5in}<{\centering}p{0.5in}<{\centering}p{0.3in}<{\centering}p{0.3in}<{\centering}}
		\midrule[1.5pt]
		& Method & \multicolumn{1}{c}{CHAOS T1} & \multicolumn{1}{c}{CHAOS T2} & \multicolumn{1}{c}{ACDC} & \multicolumn{1}{c}{LVSC} \\ 
		\cmidrule(r){1-2} 
		\cmidrule(lr){3-3} \cmidrule(lr){4-4} 
		\cmidrule(lr){5-5} \cmidrule(lr){6-6}
		
		\multirow{2}{*}{\rotatebox{90}{DSC}}
		& w/ stop-grad. & 53.8\tiny$14$ & 59.7\tiny$12$ & 82.6\tiny$04$ & 52.8\tiny$25$ \\
		& w/o stop-grad. & \underline{68.0\tiny$07$} & \underline{73.7\tiny$08$} & \underline{82.9\tiny$04$} & \underline{61.4\tiny$22$} \\
		\midrule
		\multirow{2}{*}{\rotatebox{90}{HD95}}
		& w/ stop-grad. & 26.2\tiny$18$ & 18.1\tiny$09$ & 5.4\tiny$01$ & 21.1\tiny$24$ \\
		& w/o stop-grad. & \underline{14.1\tiny$03$} & \underline{12.2\tiny$06$} & \underline{4.3\tiny$01$} & \underline{11.9\tiny$15$} \\
		\midrule[1.5pt]
	\end{tabular}
\end{table}

\begin{table}[t]
	\centering
	\caption{Comparison of different augmentation strengths.
		The best results are underlined.}
	\label{tab:strength}
	\footnotesize
	\begin{tabular}{p{0.01in}p{0.65in} | p{0.35in}<{\centering} p{0.35in}<{\centering} p{0.35in}<{\centering} p{0.35in}<{\centering}}
		\midrule[1.5pt]
		& Dataset & 1/8 & 1/4 & 1/2 & 1 \\ \midrule
		\multirow{4}{*}{\rotatebox{90}{DSC}} & CHAOS T1 & 58.2\tiny$09$ & 61.3\tiny$08$ & 63.7\tiny$09$ & \underline{68.0\tiny$07$}  \\
		& CHAOS T2 & 62.5\tiny$08$ & 66.5\tiny$07$ & 70.1\tiny$07$ & \underline{73.7\tiny$08$}  \\
		& ACDC & 82.8\tiny$04$ & 82.7\tiny$05$ & \underline{83.0\tiny$05$} & 82.9\tiny$04$  \\
		& LVSC & 54.1\tiny$22$ & 56.5\tiny$23$ & 58.4\tiny$24$ & \underline{61.4\tiny$22$}  \\
		\midrule
		\multirow{4}{*}{\rotatebox{90}{HD95}} & CHAOS T1 & 17.6\tiny$05$ & 16.4\tiny$04$ & 15.3\tiny$03$ & \underline{14.1\tiny$03$} \\
		& CHAOS T2 & 14.2\tiny$06$ & 12.9\tiny$05$ & 12.4\tiny$06$ & \underline{12.2\tiny$06$} \\
		& ACDC & 4.7\tiny$01$ & 4.6\tiny$01$ & 4.5\tiny$01$ & \underline{4.3\tiny$01$} \\
		& LVSC & 14.8\tiny$16$ & 14.4\tiny$18$ & 13.6\tiny$20$ & \underline{11.9\tiny$15$} \\
		\midrule[1.5pt]
	\end{tabular}
\end{table}

The two settings present different optimization trajectories.
With the stop-gradient, $\tilde{y}$ is treated as a constant and the network $f_\theta(\cdot)$ only receives the gradients of $f_\theta(\beta\circ\omega(x))$ in~$\mathcal{L}_{cr}$.
On the other hand, without stop-gradient, the network $f_\theta(\cdot)$ jointly receives the gradients of both $\tilde{y}$ and $f_\theta(\beta\circ\omega(x))$ in~$\mathcal{L}_{cr}$.\footnote{A derivation is given: $\nabla_\theta \mathcal{L}_{cr} = -\frac{1}{N} \sum_{i=0}^{N-1} \sum_{k=0}^{K-1} (\log y'_{ik}\nabla_\theta \tilde{y}_{ik} +  \frac{\tilde{y}_{ik}}{y'_{ik}} \nabla_\theta y'_{ik})$, where $y' = \texttt{softmax}(f_\theta(\beta\circ\omega(x)))$. 
With stop-gradient, the first term in $\nabla_\theta \mathcal{L}_{cr}$ is removed, which is the derivatives of the pseudo-mask pixel $\tilde{y}_{ik}$ with respect to the parameters $\theta$.}
As seen in Table~\ref{tab:stopgrad}, the entry without stop-gradient (ours) performs better than the one with stop-gradient.
Our explanation for this finding is that without stop-gradient, the network is jointly updated by $\mathcal{L}_{cr}$ to produce consistent $\tilde{y}$ and $f_\theta(\beta\circ\omega(x))$,
which could encourage beneficial smooth optimization via the consistency regularization.
In contrast, with stop-gradient, the network is updated by $\mathcal{L}_{cr}$ to produce solely better $f_\theta(\beta\circ\omega(x))$,
which could lead to misalignment between $\tilde{y}$ and $f_\theta(\beta\circ\omega(x))$ and undesirably impede the optimization.

\begin{figure}[t]
	\includegraphics[width=\columnwidth]{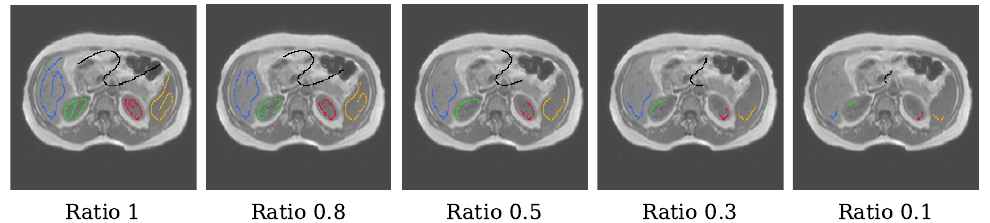}
	\centering
	\caption{Scribbles of different lengths.
		The scribbles are shorten by iteratively pruning their endpoints.
		It can be observed that scribbles gradually shrink to small line segments.}
	\label{fig:ratios}
\end{figure}

\subsection{Strengths of the Further Augmentation Operation}
\label{sec:augstrength}
In this subsection, we verify whether the further augmentation operation $\beta(\cdot)$ can influence performance.
Our assumption is that lowering the augmentation strength, which in effect contracts the neighborhood of the training set, would lead to worse performance.
To gather evidence, we gradually decrease the strength $\delta$ from 1 (the default setting) to 1/2, 1/4, and 1/8.
Note a smaller value of $\delta$ shrinks the range of a uniform distribution from which the augmentation magnitude is drawn, and
at the extreme of $\delta=0$, no further augmentation is applied.

As seen in Table~\ref{tab:strength}, decreasing the strength $\delta$ apparently deteriorates performance in both metrics on CHAOS T1\&T2 and LVSC.
These results support our assumption and suggest the important role of the further augmentation operation in our method.
However, on ACDC, while it can be observed that the HD95 results present a similar declining trend, the DSC results are quite stable.
This could be because our performance in DSC on ACDC is easily saturated, since the baseline already achieves competitive results (see Table~\ref{tab:ablation}).

\begin{figure*}[t]`
	\includegraphics[width=\textwidth]{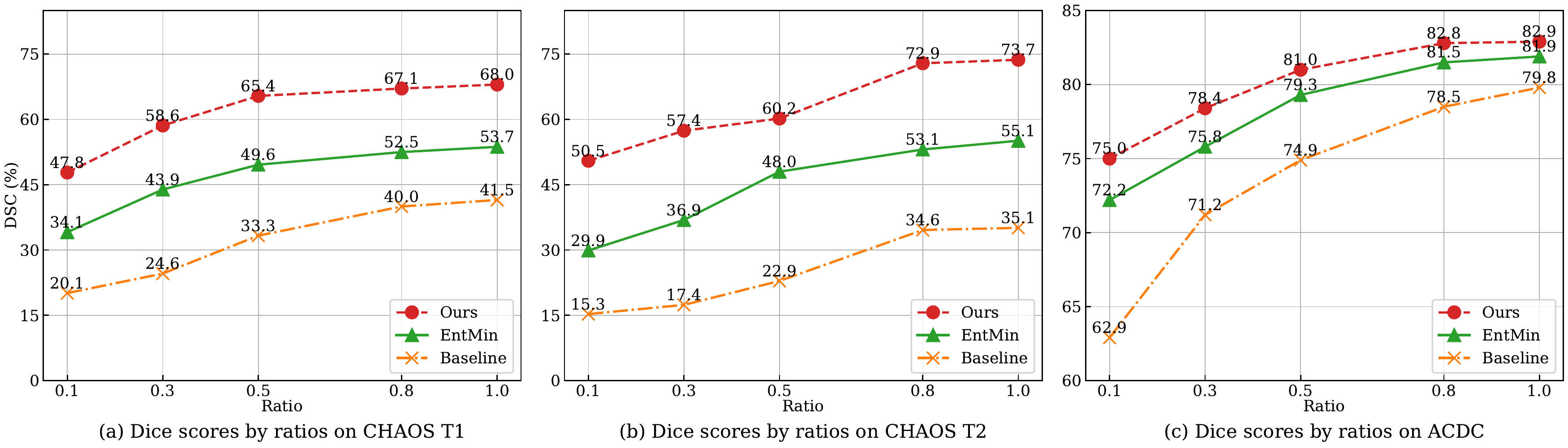}
	\centering
	\caption{Robustness to different scribble ratios. The ratio quantifies 
		the pixel number in the pruned scribbles to that in the original scribbles.}
	\label{fig:ratiocurves}
\end{figure*}

\begin{figure}[t]
	\includegraphics[width=0.98\columnwidth]{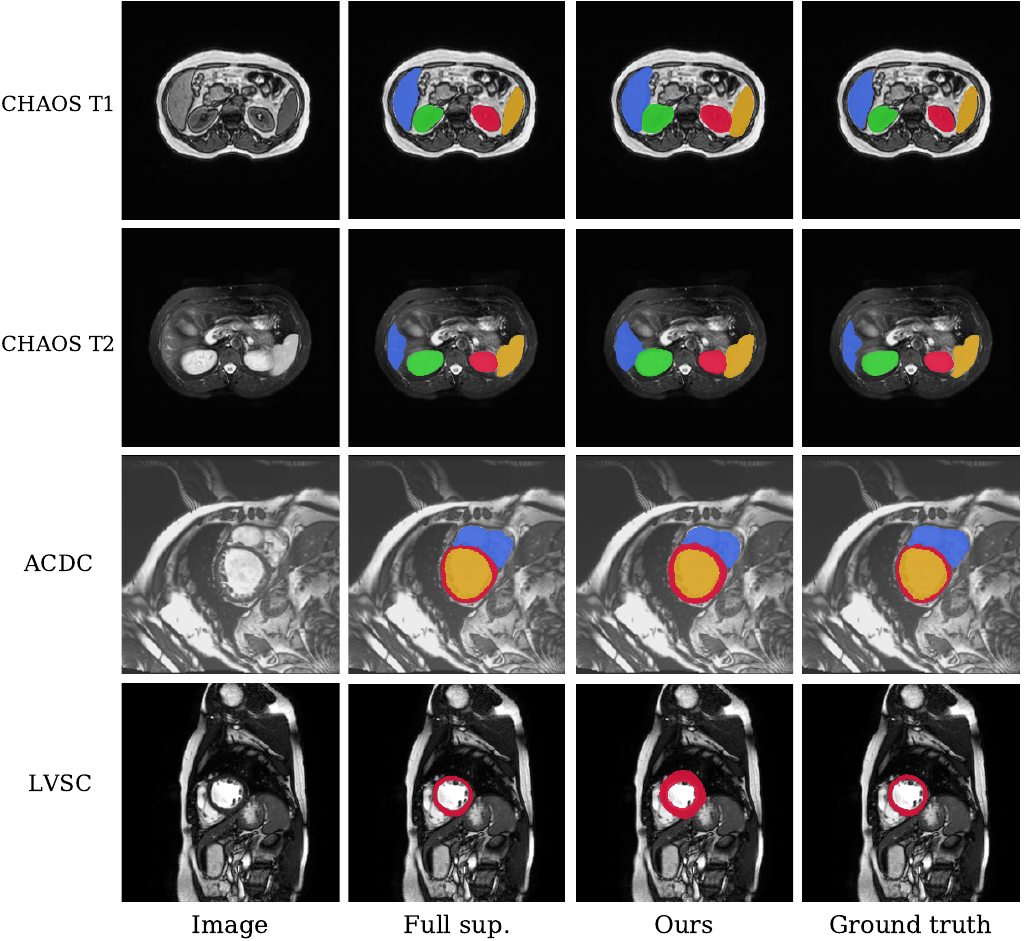}
	\centering
	\caption{Limitations of our scribble-supervised segmentation method.
		Some of our results are expanded outside the ground-truth boundaries.}
	\label{fig:lim}
\end{figure}

\subsection{Robustness to Scribble Lengths}
\label{sec:scblen}
In this subsection, we study how the number of scribble pixels covering over targets influences performance, to reassure a concern that is
high occupation of target regions by scribbles can easily produce satisfactory performance.
Specifically, we evaluate our method's robustness to scribble length.
We train a network supervised by scribble annotations of $0.1\times$, 0.3$\times$, 0.5$\times$ and 0.8$\times$original scribble pixels (Fig.~\ref{fig:ratios}).
To shorten a scribble, we detect and prune its endpoints; if the endpoint does not exist, we prune a random scribble pixel.
This process is iterated until a length requirement (e.g., 0.8$\times$original scribble pixels) is met.
Experiments are conducted on CHAOS T1\&T2 and ACDC with real-world scribble annotations.
We compare our method with both the baseline and EntMin (seen as the improved baseline).

As illustrated in Fig.~\ref{fig:ratiocurves}, our method consistently outperforms the baseline and EntMin under different scribble ratios.
Furthermore, although it can be observed that a smaller ratio produce a lower DSC result,
this trend is stable at certain intervals.
To be specific, the DSC result drops by a trivial margin from 1.0$\times$ to 0.5$\times$scribble pixels on CHAOS T1, 
and from 1.0$\times$ to 0.8$\times$scribble pixels on CHAOS T2 and ACDC.
This finding suggests that, within certain worse variations in scribble quality, our method can still achieve good performance.

\subsection{Limitations}
\label{sec:lim}
While PacingPseudo achieves overall satisfactory results, there still remain limitations.
Some segmentation results tend to be expanded outside the ground-truth boundaries.
As seen in Fig.~\ref{fig:lim}, on CHAOS T1\&T2, while the segmentations of the full supervision distinguish different organs clearly, 
our results fail to provide accurate boundaries in the adjacent regions between different organs (e.g., the liver and right kidney).
On ACDC and LVSC, while the thin myocardium is well delineated by the full supervision, our method generates much thicker segmentation regions.
We conjecture these issues could be due to the inherent limitations of the insufficient supervision from the scribble annotations.

One possible solution to these issues is constraining unlabeled pixels' predictions based on low-level image information.
Dorent et al.~\citep{dorent2020scribble} and Meng et al.~\citep{tang2018regularized} propose regularizers based on DCRFs. 
Kolesnikov et al.~\citep{kolesnikov2016seed} devise a loss function that compels a network to produce predictions analogous to DCRF-processed segmentations.
These methods enable neural networks to learn low-level image information. PacingPseudo could potentially benefit from learning low-level image information and 
produce more precise boundaries.

\section{Conclusion}
We propose an effective non-iterative method for scribble-supervised segmentation in medical imaging.
Our method attains a non-iterative training paradigm by means of a weight-sharing siamese architecture, 
wherein pseudo-masks reinforce network learning during training.
Inspired by insights in consistency training, our designs to boost performance include the entropy regularization 
to obtain high-confidence pseudo-masks and the distorted augmentations to create discrepancy for consistency regularization.
Besides, to complement scarce labeled pixels, we devise a memory bank that introduces extra cross-image ensemble features. 
Experimental results show that our method can bridge the gap between scribble-supervised and fully-supervised segmentation to some extent, and proves robust even to some undesirable scribbles.
Thanks to its simplicity, the proposed method has the potential to be extended to 3D by using volumetric scribble-annotating approaches.






\bibliographystyle{model2-names} \biboptions{authoryear}


%
%
%

\end{document}